\documentclass[authoryear]{elsarticle}

\usepackage{lineno,hyperref}
\modulolinenumbers[5]
\let\linenumbers\nolinenumbers\nolinenumbers
\journal{Computer Speech and Language}

\usepackage{setspace,url}
\PassOptionsToPackage{hyphens}{url}
\usepackage{lineno,multirow,cite}
\usepackage{epsfig,amssymb}
\usepackage[utf8]{inputenc}
\usepackage{amsmath,graphicx}
\usepackage{amssymb,epsfig,url,mathrsfs} \usepackage{algorithm,algorithmic}
\usepackage{xcolor}
\usepackage{multicol}
\usepackage{enumitem}
\usepackage{numprint}
\npthousandsep{,}
\usepackage{graphicx}
\usepackage{array, makecell}
\usepackage{verbatim}
\usepackage{subcaption}

\biboptions{authoryear}
\renewcommand{\cite}{\citep}

\usepackage{dirtytalk}
\usepackage{breakcites}
\usepackage{tabularx}
\usepackage{tikz}
\usepackage{color, colortbl}
\usepackage{bm}
\usepackage{todonotes} 
\usepackage{changebar} 
\usepackage{svg}
\definecolor{chestnut}{rgb}{0.8, 0.36, 0.36}
\definecolor{frenchblue}{rgb}{0.0, 0.45, 0.73}
\definecolor{babyblue}{rgb}{0.54, 0.81, 0.94}
\definecolor{beaublue}{rgb}{0.74, 0.83, 0.9}
\definecolor{bubblegum}{rgb}{0.99, 0.76, 0.8}
\definecolor{gainsboro}{rgb}{0.86, 0.86, 0.86}
\definecolor{amethyst}{rgb}{0.6, 0.4, 0.8}
\definecolor{mygray}{gray}{0.85}
	\definecolor{aq}{rgb}{0.5, 1.0, 0.83}

\setlist[itemize]{leftmargin=4.5mm}
\usepackage{xr}

\makeatletter
\newcommand*{\addFileDependency}[1]{
  \typeout{(#1)}
  \@addtofilelist{#1}
  \IfFileExists{#1}{}{\typeout{No file #1.}}
}
\makeatother

\makeatletter

\usepackage{comment}
\let\wfs@comment@comment\comment
\let\comment\@undefined

\usepackage[final]{changes}
\let\wfs@changes@comment\comment
\let\comment\@undefined

\newcommand\comment{%
    \ifthenelse{\equal{\@currenvir}{comment}}
    {\wfs@comment@comment}
    {\wfs@changes@comment}%
}

\makeatother

\usepackage{cancel}
\usepackage{changes}
\usepackage{lipsum}

\definechangesauthor[name=revision, color=blue]{rev}
\definechangesauthor[name=comment, color=orange]{com}

\begin{document}

\begin{frontmatter}

\title{The VoicePrivacy 2020 Challenge:  Results and findings}

\author[mymainaddress1]{Natalia Tomashenko\corref{mycorrespondingauthor}}
\cortext[mycorrespondingauthor]{Corresponding author}
\ead{natalia.tomashenko@univ-avignon.fr}
\author[mysecondaryaddress2]{Xin Wang}
\author[mysecondaryaddress3]{Emmanuel Vincent}
\author[mysecondaryaddress4]{Jose Patino}
\author[mysecondaryaddress6]{Brij Mohan Lal Srivastava}
\author[mymainaddress1]{Paul-Gauthier Noé}
\author[mysecondaryaddress4]{Andreas Nautsch}
\author[mysecondaryaddress4]{Nicholas Evans}
\author[mysecondaryaddress2,mysecondaryaddress5]{Junichi Yamagishi}
\author[mysecondaryaddress7]{Benjamin O’Brien}
\author[mymainaddress1]{Anaïs Chanclu}
\author[mymainaddress1]{Jean-François Bonastre}
\author[mysecondaryaddress4]{Massimiliano Todisco}
\author[mysecondaryaddress6]{Mohamed Maouche}

\address[mymainaddress1]{LIA, University of Avignon, Avignon, France}
\address[mysecondaryaddress2]{National Institute of Informatics (NII), Tokyo, Japan}
\address[mysecondaryaddress3]{Université de Lorraine, CNRS, Inria, LORIA, France}
\address[mysecondaryaddress4]{Audio Security and Privacy Group, EURECOM, France}
\address[mysecondaryaddress5]{University of Edinburgh, UK}
\address[mysecondaryaddress6]{Inria, France}
\address[mysecondaryaddress7]{LPL, Aix-Marseille University, France}

\begin{abstract}
This paper presents the results  and analyses stemming from the first VoicePrivacy 2020 Challenge
which focuses on developing anonymization solutions for speech technology. We provide a
systematic overview of the challenge design with an analysis of submitted systems and evaluation results. In particular,  
we describe the voice anonymization task and datasets used for system development and evaluation.
Also, we present different attack models and the associated objective and subjective evaluation metrics. 
We introduce two anonymization baselines and provide a summary description of the anonymization systems developed by the challenge participants.
We report
objective and subjective evaluation results
for baseline and submitted systems. 
In addition, we present experimental results for 
alternative privacy metrics and attack models developed as a part of the post-evaluation analysis.
Finally, we summarise our insights and observations that will influence the design of the next VoicePrivacy challenge edition and some directions for future voice anonymization research.
\end{abstract}


\end{frontmatter}

\linenumbers

\section{Introduction}

Due to the growing demand for privacy preservation  in the recent years,
privacy-preserving data processing has become an active research area.
One reason for this is the European general data protection regulation (GDPR)  in  the European Union (EU) law and similar regulations
 in  national laws of many countries outside the EU concerning the implementation of the data protection principles when treating, transferring or storing personal data. 
 
 Although a legal  definition of privacy  is missing \cite{nautsch2019gdpr},
 speech data contains a lot of personal information that can be disclosed by listening or by automated systems \cite{Nautsch-PreservingPrivacySpeech-CSL-2019}. This includes, e.g., age, gender, ethnic origin, geographical background, health or emotional state, political orientations, and religious beliefs.
 Speaker recognition systems can also reveal the speaker's identity. 
 Therefore, the increased interest in
developing privacy preservation solutions for speech technology is not surprising.
 This motivated the launching of the VoicePrivacy initiative \cite{tomashenko2020introducing}. This  initiative aims to
bring together a new community of researchers,  engineers and privacy professionals
in order to  formulate the tasks of interest, develop  evaluation methodologies, and
benchmark new solutions through a series of challenges. The first VoicePrivacy challenge\footnote{\url{https://www.voiceprivacychallenge.org/}} was organized  as a part of this initiative \cite{tomashenko2020introducing, tomashenkovoiceprivacy}.

 Existing approaches to privacy preservation for speech  can be broadly classified into: obfuscation, encryption, distributed learning, or anonymization. 
 \added[id=rev, comment=1.4]{Obfuscation methods \cite{cohen2019voice,gontier2020privacy} suppress or modify the speech signal to the point where no information about it can be recovered. Encryption methods \cite{pathak2013privacy,brasser2018voiceguard,zhang2019encrypted} support computation upon data in the encrypted domain, however they significantly increase the computational complexity.
 Decentralized or federated learning methods learn models from distributed data without accessing it directly \cite{leroy2019federated}, however the derived data used for learning (e.g., model gradients) may still leak information about the original data \cite{tomashenko2021privacy,mdhaffar2021retrieving}.
 Note also that the latter two categories of approaches are incompatible with using the data for supervised machine learning purposes, which requires third-party annotators to access the data in non-encrypted form.}

\emph{Anonymization} refers to the goal of suppressing personally identifiable information in the speech signal, leaving other attributes intact.
\added{In contrast to the above approaches, it allows the data to be used for supervised machine learning purposes and it can easily be integrated within existing systems.}
Note, that in the legal community, the term ``anonymization'' means that this goal has been achieved. Here, it refers to the task to be addressed, even when the method being evaluated has failed.
\added[id=rev, comment=2.2]{Anonymization requires altering not only the speaker's voice, but also other traits and states, words in the spoken contents, and sounds in the background which, when considered in combination with each other and possibly with external data, may reveal the speaker's identity}.

\added[id=rev, comment=2.2]{As a first step towards this goal, the VoicePrivacy 2020 Challenge focuses on \textit{voice anonymization}, that is the task of altering the speaker's voice to hide their identity to the greatest possible extent, while leaving all other speech attributes (traits, states, and spoken contents) intact.
Approaches to voice anonymization} include 
noise addition \cite{hashimoto2016privacy}, speech transformation \cite{qian2017voicemask, patino2020speaker}, voice conversion \cite{fang2019speaker,han2020voice,srivastava2020baseline}, and disentangled representation learning \cite{srivastava2019privacy,aloufi2020privacy}.

Despite the appeal of voice anonymization, the level of privacy protection offered by these solutions is unclear and not meaningful because there is no formal definition of the task and no formal attack model, and there are no common datasets, protocols and metrics. 
The VoicePrivacy 2020 Challenge aims to address all of these concerns.

The paper is structured as follows.
The challenge design, including the description of the anonymization task, attack models, datasets, objective and subjective evaluation methodologies with the corresponding privacy and utility metrics,  is presented in Section~\ref{sec:challenge_design}.
The overview of the  baseline  and  submitted systems  is provided in Sections~\ref{sec:system_description}.
Objective and subjective evaluation results and their comparison and analysis are presented in Section~\ref{sec:results}.
We conclude and discuss future directions in Section~\ref{sec:conclusions}.

\section{Challenge design}\label{sec:challenge_design}

In this section, we present an overview of the official challenge setup: anonymization task, corresponding attack models selected for the challenge, data and evaluation methodology. Also we present an additional attack model developed as part of the post-evaluation analysis \cite{tomashenko2020posteval}.

\subsection{Anonymization task and attack models}
\label{sec:task}

Privacy preservation is formulated as a game between \emph{users} who 
share\footnote{\added{This data may be shared with selected individuals, with a company providing a service, with a public cloud provider, with the general public (open data), etc.. Attackers may include employees or subcontractors of these companies, hackers who get access to the cloud storage, or simply other individuals who browse the open data.}} some data and \emph{attackers} who access this data or data derived from it and wish to infer information about the users \cite{qian2018towards,srivastava2019evaluating,tomashenko2020introducing}. To protect their privacy, the users \added[id=rev, comment=1.2]{share} data that contain as little personal information as possible while allowing one or more downstream goals to be achieved. To infer personal information, the attackers may use additional prior knowledge.

Focusing on speech data, a given privacy preservation scenario is specified by:
(i)~the nature of the data: waveform, features, etc., (ii)~the information seen as personal: speaker identity, traits, spoken contents, etc., (iii)~the downstream goal(s): human communication, automated processing, model training, etc., (iv)~
the data accessed by the attackers: one or more utterances, derived data or model, etc., (v)~the attackers' prior knowledge: previously \added[id=rev, comment=1.2]{shared} data, privacy preservation method applied, etc.
Different specifications lead to different privacy preservation methods from the users' point of view and different attacks from the attackers' point of view.

\begin{figure}[htbp]
\centering\includegraphics[width=\textwidth]{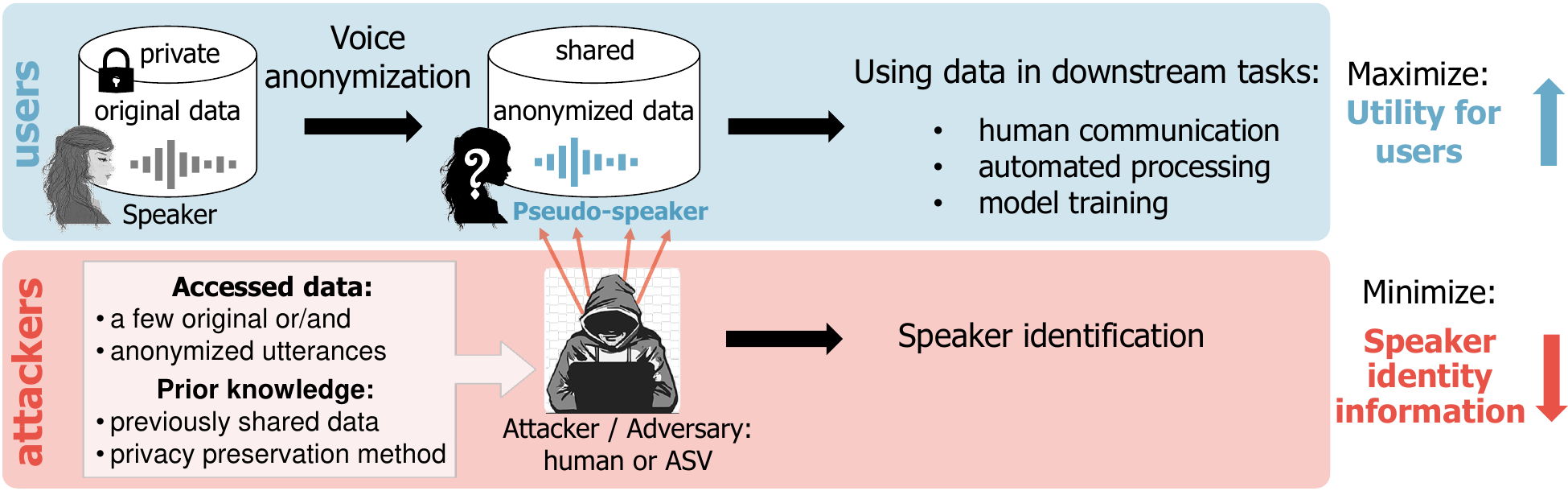}
\caption{Example of a privacy preservation scenario   as a game between  \emph{users} and \emph{attackers} in the case where speaker identity is considered as personal information to be protected.}
\label{fig:game}
\end{figure}

\added[id=rev,comment=1.5]{Here, we consider the scenario illustrated in Figure~\ref{fig:game} where speakers want to hide their identity to the greatest possible extent while allowing the desired downstream goals to be achieved, while attackers want to identify the speakers from their utterances.}

\subsubsection{Anonymization task}\label{subsec:anon_task}
The \added[id=rev, comment=1.2]{sentences shared by the users} are called \emph{trial} utterances.\footnote{\added{The terms \textit{trial} and \textit{enrollment} are borrowed from the speaker verification literature, where they refer respectively to a speech signal uttered by a speaker willing to be authenticated and a speech signal (or a model) associated with the claimed identity. Although anonymization is a different task (there is no speaker willing to be authenticated), these terms are used here due to the high similarity between the evaluation protocols for these two tasks.}} In order to hide his/her identity, each user passes these utterances through a voice anonymization system \added{prior to sharing}. The resulting utterances sound as if they were uttered by another speaker, which we call \emph{pseudo-speaker} since it may be an artificial voice not corresponding to any real speaker.

The task of challenge participants is to develop this anonymization system. 
It should: (a)~output a speech waveform, (b)~hide speaker identity, (c)~leave other speech characteristics unchanged, (d)~ensure that all trial utterances from a given speaker are uttered by the same pseudo-speaker, while trial utterances from different speakers are uttered by different pseudo-speakers.

The requirement (c) promotes the achievement of all possible downstream goals to the best possible extent. In practice, we restrict ourselves to a few goals corresponding to two use cases: ASR training and/or decoding, and multi-party human conversations.
The  requirement (d) corresponds to the latter goal and is motivated by the fact that, in a multi-party human conversation, the anonymized voices of all speakers must sound natural, be distinguishable from each other, and cannot change over time. 
\replaced[id=rev, comment=2.1]{The achievement of these goals}{This was assessed via a new, specifically designed metric --- \emph{gain of voice distinctiveness}.}
is assessed via a range of \emph{utility} metrics.

\subsubsection{Attack models}
\label{subsec:attack_model}

\added[id=rev, comment=1.15]{For each speaker of interest, the attacker is assumed to have access to one or more utterances spoken by that speaker. These utterances
 may or may not have been
anonymized and are called \textit{enrollment} utterances.}

\added{In this work, the}
attackers have access to: (a) one or more anonymized trial utterances, (b) possibly, original or anonymized enrollment utterances for each speaker. \deleted[id=rev, comment=2.1]{They do not have access to the anonymization system applied by the user.} 
The protection of identity information is assessed via \emph{privacy} metrics, including objective speaker verifiability and subjective speaker verifiability and linkability. These metrics assume different attack models.

The objective speaker verifiability metrics (Section \ref{sec:obj_eval_metr}) assume that the attacker has access to a single anonymized trial utterance and several enrollment utterances. 
    Two sets of metrics were computed, corresponding to the two attack models when the enrollment utterances are original or they have been anonymized \replaced[id=rev, comment=2.1]{by the user or the attacker.}{In the latter case, we assume that the utterances have been anonymized in the same way as the trial data using the same anonymization system, i.e., all enrollment utterances from a given speaker are converted into the same pseudo-speaker, and enrollment utterances from different speakers are converted into different pseudo-speakers. We also assume that the pseudo-speaker corresponding to a given speaker in the enrollment set is different from the pseudo-speaker corresponding to that same speaker in the trial set.}
    In the \textit{post-evaluation} stage, we considered \deleted{alternative anonymization procedures corresponding to} a stronger attack model where
    attackers also have access to anonymized training data and can retrain an automatic speaker verification system using this data. \deleted[id=rev, comment=2.1]{We assume that the training,  enrollment and trial data have been anonymized using the same  system with different corresponding pseudo-speakers.}
  
   For the subjective evaluation (Section \ref{sec:subj_eval_meth}), two situations are considered. The speaker verifiability metric assumes that the attacker has access to a single anonymized trial utterance and a single original enrollment utterance, while the speaker linkability metric assumes that the attacker has access to  several original and anonymized trial utterances.

\subsection{Datasets}\label{sec:data}
Several publicly available corpora are used for the training, development and evaluation of voice anonymization systems.

\paragraph{Training set}
The training set comprises the \numprint{2800}~h \textit{VoxCeleb-1,2} corpus \cite{nagrani2017voxceleb,chung2018voxceleb2} and 600~h subsets of the \textit{{LibriSpeech}} \cite{panayotov2015librispeech} and \textit{LibriTTS} \cite{zen2019libritts} corpora. \added[id=rev, comment=2.3]{These corpora are among the largest and the most widely used for speaker verification, ASR, and speech synthesis, respectively, hence they are natural choices for training voice anonymization systems which must extract speaker identity and phonetic information and resynthesize a speech signal which hides the former and preserves the latter.} The selected subsets are detailed in Table~\ref{tab:data} (top).

\begin{table}[htbp]
\centering
  \caption{Number of speakers and utterances in the VoicePrivacy 2020 training, development, and evaluation sets.}\label{tab:data}
 \resizebox{0.85\textwidth}{!}{
  \centering
  \begin{tabular}{|c|l|l|r|r|r|r|}
\hline
 \multicolumn{3}{|l|}{\textbf{Subset}} &  \textbf{Female} & \textbf{Male} & \textbf{Total} & \textbf{\#Utter.}  \\ \hline \hline
\multirow{5}{*}{\rotatebox{90}{Training~}} & \multicolumn{2}{l|}{VoxCeleb-1,2} & \numprint{2912} & \numprint{4451} & \numprint{7363} & \numprint{1281762} \\ \cline{2-7}
& \multicolumn{2}{l|}{LibriSpeech train-clean-100} & 125 & 126 & 251	& \numprint{28539} \\ \cline{2-7}
& \multicolumn{2}{l|}{LibriSpeech train-other-500} & 564 & 602 & \numprint{1166} & \numprint{148688}	\\ \cline{2-7}
& \multicolumn{2}{l|}{LibriTTS train-clean-100} & 123 & 124 & 247 & \numprint{33236} \\ \cline{2-7}
& \multicolumn{2}{l|}{LibriTTS train-other-500} & 560 & 600 & \numprint{1160} & \numprint{205044} \\ \hline\hline
\multirow{5}{*}{\rotatebox{90}{~ Development }} & LibriSpeech & Enrollment & 15 & 14 & 29 & 343\\ \cline{3-7}
& dev-clean & Trial & 20 & 20 & 40 & \numprint{1978}\\ \cline{2-7}
& & Enrollment & & & & 600 \\  \cline{3-3}\cline{7-7}
& VCTK-dev & Trial (common) & 15 & 15 & 30 & 695\\  \cline{3-3}\cline{7-7}
& & Trial (different) & & & & \numprint{10677} \\  \hline\hline
\multirow{5}{*}{\rotatebox{90}{Evaluation~}} & LibriSpeech & Enrollment & 16 & 13 & 29 & 438\\ \cline{3-7}
& test-clean & Trial & 20 & 20 & 40 & \numprint{1496}\\ \cline{2-7}
& & Enrollment & & & & 600 \\  \cline{3-3}\cline{7-7}
& VCTK-test & Trial (common) & 15 & 15 & 30 & 700 \\ \cline{3-3}\cline{7-7}
& & Trial (different) & & & & \numprint{10748} \\ \hline
\end{tabular}}
\end{table}
\normalsize

\paragraph{Development set}
The development set involves \textit{LibriSpeech dev-clean} and a subset of the VCTK corpus \cite{yamagishi2019cstr}, denoted  \textit{VCTK-dev} (see Table~\ref{tab:data}, middle). With the above attack models in mind, we split them into trial and enrollment subsets. For \textit{LibriSpeech dev-clean}, the speakers in the enrollment set are a subset of those in the trial set. \added[id=rev, comment=2.3]{This corpus is meant for objective ASR performance evaluation.} For \textit{VCTK-dev}, we use the same speakers for enrollment and trial and we consider two trial subsets: \textit{common} and \textit{different}. The \textit{common} subset comprises utterances $\#1-24$ in the VCTK corpus that are identical for all speakers. This is meant for subjective evaluation of speaker verifiability/linkability in a text-dependent manner. The enrollment and \textit{different} subsets comprises distinct utterances for all speakers.

\paragraph{Evaluation set} Similarly, the evaluation set comprises \textit{LibriSpeech test-clean} and a subset of VCTK called \textit{VCTK-test} (see Table~\ref{tab:data}, bottom).

\subsection{Utility and privacy metrics}
\label{sec:metrics}

We consider objective and subjective privacy metrics to assess speaker re-identification and linkability. We also propose objective and subjective utility metrics to assess the fulfillment of the user goals specified in Section \ref{sec:task}. \deleted{Specifically, we consider ASR performance using a model trained on original data and subjective speech intelligibility and naturalness.}

\subsubsection{Objective metrics}
\label{sec:obj_eval_metr}

For objective evaluation of anonymization performance, two systems were trained to assess the following characteristics: 
(1) speaker verifiability and  (2)  ability of the anonymization system to  
  preserve linguistic information in the anonymized speech.
The first system, denoted  \textrm{$ASV_\text{eval}$}, is an automatic speaker verification (ASV) system
based on x-vector speaker embeddings and probabilistic linear discriminant analysis (PLDA) \cite{snyder2018x}, which outputs a log-likelihood ratio (LLR) score.
The second system, denoted  \textrm{$ASR_\text{eval}$}, is an automatic speech recognition (ASR) system which outputs a word \added{sequence}.
Both \textrm{$ASR_\text{eval}$} and \textrm{$ASV_\text{eval}$} were trained on the \textit{LibriSpeech-train-clean-360} dataset using the Kaldi speech recognition toolkit \cite{povey2011kaldi}. 
These two models were used in the official  challenge setup \cite{tomashenko2020introducing}.
In addition, for post-evaluation analysis, we trained ASV and ASR systems  on anonymized speech data. Both models, denoted  $ASV_\text{eval}^\text{anon}$ and $ASR_\text{eval}^\text{anon}$, were trained in the same way as $ASV_\text{eval}$ and $ASR_\text{eval}$, respectively.\footnote{Scripts for training \textrm{$ASR_\text{eval}$} and \textrm{$ASV_\text{eval}$} and for evaluation are provided at \url{https://github.com/Voice-Privacy-Challenge/Voice-Privacy-Challenge-2020}.}

\added[id=rev, comment=1.6 \& 2.3]{For objective utility evaluation, the official challenge setup relies on the ubiquitous word error rate (WER) metric. The post-evaluation phase also considers the \emph{gain of voice distinctiveness} metric of \citet{noe2020speech}, which accounts for the additional requirement that the anonymized voices of all speakers must be distinguishable from each other.}

\begin{figure}[t!]
\captionsetup{singlelinecheck=off}
\centering\includegraphics[width=115mm]{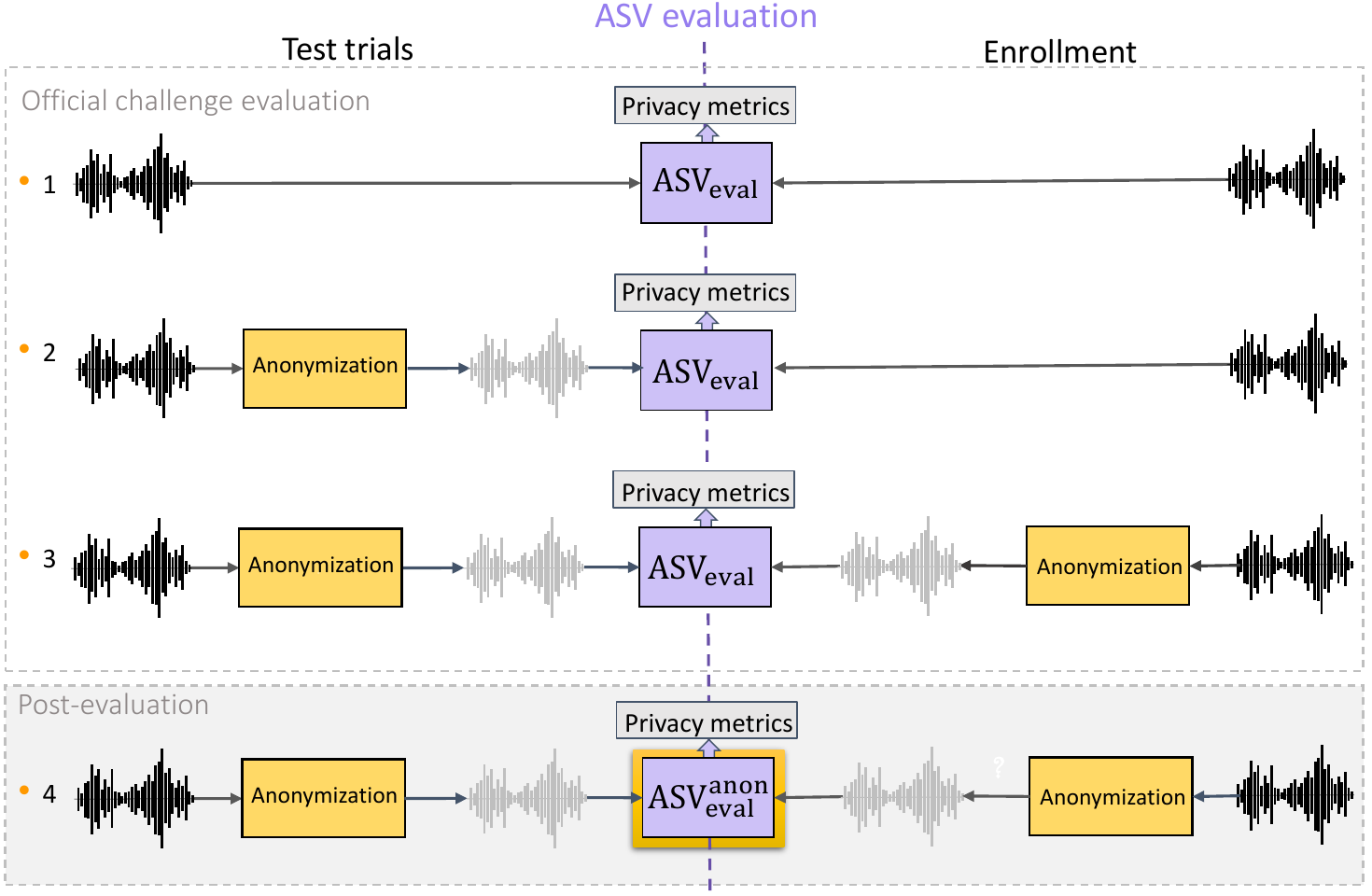}
\caption[foo bar]{ASV evaluation for the official challenge setup using $ASV_\text{eval}$ trained on original data is performed for three cases:
    (1)\textit{\textcolor{amethyst}{Unprotected}}: 
    \added{original enrollment and trial;} 
    (2) \textit{\textcolor{amethyst}{Ignorant attacker (oa)}}: \added{original enrollment and anonymized trial;}
    (3) \textit{\textcolor{amethyst}{Lazy-informed attacker~(aa)}}: \added{anonymized  enrollment and trial.}
  ASV evaluation for the post-evaluation analysis is performed using $ASV_\text{eval}^\text{anon}$ trained on anonymized data  for case
    (4) \textit{\textcolor{amethyst}{Semi-informed attacker~(aa)}}:  \added{anonymized enrollment and trial.}
}
\label{fig:asv-eval}
\end{figure}

 \added{For objective privacy evaluation}, three \added[id=rev, comment=2.3]{well-established speaker verification} metrics are computed in the official challenge setup: the \emph{equal error rate (EER)} and the \emph{log-likelihood ratio (LLR)} based costs $C_\text{llr}$ and $C^\text{min}_\text{llr}$.
As seen in Figure~\ref{fig:asv-eval}, these metrics are computed for 4 evaluation scenarios corresponding to different  types of attacks depending on the amount of the attackers' knowledge. Following the terminology of \citet{srivastava2021}, we consider the following conditions.
\begin{enumerate}\setlength\itemsep{0.05em}
\item \textit{Unprotected}: no anonymization is performed by users; attackers have access to original trial and enrollment data.
\item \textit{Ignorant attacker (oa)}:
original enrollment and anonymized trial data are used for evaluation. We refer to this scenario as \textit{(\textbf{o}riginal, \textbf{a}nonymized)} or \textit{oa} in short.
Users anonymize their trial data, but attackers are unaware of it, hence they use original data for enrollment.
\item \textit{Lazy-informed (aa)}
anonymized enrollment and anonymized trial data are used for evaluation. We refer to this scenario as \textit{(\textbf{a}nonymized, \textbf{a}nonymized)} or \textit{aa} in short.
This scenario reflects the situation when the enrollment data are anonymized data produced by users, who are assumed to use the same anonymization system but different pseudo-speakers from their trial data.\footnote{\added{The scenario when enrollment data has been anonymized using the same pseudo-speakers as the trial data is of little practical interest. Indeed, prior work has shown that attackers who have access to such data can identify the speakers as well as if the data had not been anonymized \citep{srivastava2019evaluating}. Users are therefore required to pick a different, random pseudo-speaker for each conversation.}}
\added{While it is unlikely that attackers have access to anonymized data} %
\added{with explicit speaker identities, they may infer the identities of a subset of the data from the spoken contents and subsequently use this data as enrollment data. This scenario also reflects the alternative situation when attackers have access to original enrollment data and anonymize them using the same system (which is assumed to be publicly available) so that they become more similar to the anonymized trial data. Here again, the data is anonymized using a different pseudo-speaker, since attackers do not know which pseudo-speaker was picked by each user. Hence, both situations result in the same attack model.}

\item \textit{Semi-informed (aa with the model retrained on anonymized data)}: attackers have the same knowledge as in the previous case (the anonymization system, but not the 
\added[id=rev, comment=2.1]{pseudo-speaker picked by each speaker}) and, in addition to this, they
\added[id=rev, comment=2.1]{anonymize the training set for the $ASV_\text{eval}$ model using the same anonymization system with different pseudo-speakers and re-train it} on this data. 
 These attackers are \deleted{supposed to be} the strongest ones among the  considered in this paper. 
 This evaluation scenario 
\added[id=rev, comment=1.8]{is part of}
 the post-evaluation stage.
\end{enumerate}

The number of \added{same-speaker and different-speaker} trials in the development and evaluation datasets is given in Table \ref{tab:trials}. 
\added{In addition to the EER, $C_\text{llr}$, and $C^\text{min}_\text{llr}$, the post-evaluation phase considers one more privacy metric, namely the \emph{de-identification} metric of \citet{noe2020speech} which assesses how different each pseudo-speaker is from the original speaker. Note that, although this metric provides useful additional information, it does not directly match the requirements set in Section \ref{sec:task}. Indeed, the requirement that the original speaker cannot be identified from the anonymized signal does not imply that the pseudo-speaker's voice must be maximally different.}

\begin{table}[htbp]
  \caption{Number of speaker verification trials.}\label{tab:trials}
  \renewcommand{\tabcolsep}{0.13cm}
  \centering
   \resizebox{0.87\textwidth}{!}{
  \begin{tabular}{|l|l|l|r|r|r|}
\hline
 \multicolumn{2}{|l|}{\textbf{Subset}} & \textbf{Trials} &  \textbf{Female} & \textbf{Male} & \textbf{Total}  \\ \hline \hline
\multirow{6}{*}{\rotatebox{90}{Development~}} & LibriSpeech & Same-speaker & 704 & 644 & \numprint{1348} \\ \cline{3-6}
 & dev-clean & Different-speaker	& \numprint{14566} & \numprint{12796} &	\numprint{27362} \\ \cline{2-6}
 & \multirow{4}{*}{VCTK-dev} & Same-speaker (common) & \numprint{344} & \numprint{351} & \numprint{695} \\ \cline{3-6}
 & & Same-speaker (different) & \numprint{1781}	& \numprint{2015} & \numprint{3796} \\  \cline{3-6}
 & & Different-speaker (common) & \numprint{4810} &	\numprint{4911} & \numprint{9721} \\ \cline{3-6}
 & & Different-speaker (different) & \numprint{13219} & \numprint{12985} & \numprint{26204} \\ \hline\hline
\multirow{6}{*}{\rotatebox{90}{Evaluation~}} & LibriSpeech & Same-speaker & 548 & 449	& \numprint{997} \\ \cline{3-6}
 & test-clean & Different-speaker & \numprint{11196} & \numprint{9457} &	\numprint{20653} \\ \cline{2-6}
 & \multirow{4}{*}{VCTK-test} & Same-speaker (common) & \numprint{346} & \numprint{354} & \numprint{700} \\ \cline{3-6}
 & & Same-speaker (different) & \numprint{1944} & \numprint{1742} & \numprint{3686} \\  \cline{3-6}
 & & Different-speaker (common) & \numprint{4838} & \numprint{4952} & \numprint{9790} \\ \cline{3-6}
 & & Different-speaker (different) & \numprint{13056} &	\numprint{13258} &\numprint{26314} \\ \hline
  \end{tabular}}
\end{table}
\normalsize

The objective evaluation metrics for privacy and utility are listed below.

\paragraph{\textbf{Equal error rate (EER)}}\label{subsec:eer}
Denoting by $P_\text{fa}(\theta)$ and $P_\text{miss}(\theta)$ the false alarm and miss rates at threshold~$\theta$, the EER corresponds to the threshold $\theta_\text{EER}$ at which the two detection error rates are equal, i.e., $\text{EER}=P_\text{fa}(\theta_\text{EER})=P_\text{miss}(\theta_\text{EER})$.

\paragraph{\textbf{{Log-likelihood-ratio cost function ($C_\text{llr}~\text{and}~\text{C}^\text{min}_\text{llr}$)}}}\label{subsec:cllr}

 $C_\text{llr}$ is computed from PLDA scores as defined by \citet{brummer2006application} and \citet{ramos2008cross}.
It can be decomposed into a discrimination loss ($C^\text{min}_\text{llr}$) and a calibration loss ($C_\text{llr}-C^\text{min}_\text{llr}$). $C^\text{min}_\text{llr}$ is estimated by optimal calibration using monotonic transformation of the scores to their empirical LLR values.

\paragraph{\textbf{De-identification and gain of voice distinctiveness}}\label{subsec:similar_matrix}
To visualize anonymization performance  across different speakers in a dataset,  voice similarity matrices have been proposed by \citet{noe2020speech}.
A voice similarity matrix $M=( M(i,j))_{1 \le i \le N,1 \le j \le N}$ is defined for a set of $N$ speakers using similarity values $ M(i,j)$ computed for speakers $i$ and $j$ as follows:
\begin{equation}
\small
     M(i,j) = \mathrm{sigmoid}\left({\frac{1}{n_{i}n_{j} }
    \displaystyle\sum_{\substack{1 \le k \le n_{i} \text{ and } 1 \le l \le n_{j} \\ k\neq l \text{ if } i=j } }{\text{LLR}(x^{(i)}_{k},x^{(j)}_{l})}}\right)
    \label{equ:: M}
\end{equation}
where $\text{LLR}(x^{(i)}_{k},x^{(j)}_{l})$ is the log-likelihood-ratio obtained by comparing the $k$-th segment from the $i$-th speaker with the $l$-th segment from the $j$-th speaker, and $n_{i}$ and $n_j$ are the numbers of segments for these speakers. 
Three matrices are computed: $M_\text{oo}$ on original data, $M_\text{aa}$ on anonymized data,
   and  $M_\text{oa}$ on original and anonymized data. For computing the entries $M(i,j)$ of $M_\text{oa}$, we use original data for speaker $i$ and anonymized data for speaker $j$.

  Using voice similarity matrices, two additional metrics can be computed: de-identification (DeID) and gain of voice distinctiveness ($G_{\text{VD}}$) \cite{noe2020speech}.
They are computed based on the ratio of diagonal dominance for two pairs of matrices: \{$M_\text{oa}$, $M_\text{oo}$\} or \{$M_\text{oo}$, $M_\text{oo}$\}, respectively.
The diagonal dominance $D_\text{diag}(M)$ is defined as the absolute difference between the mean values of  diagonal and  off-diagonal elements:
\begin{equation}
\small
    D_{\text{diag}}(M)\hspace{-0.7mm}=\hspace{-0.7mm}\displaystyle
    \Bigg|
    \sum_{1\leq i \leq N} \frac{ M(i,i)}{N}
    \displaystyle
    - 
    \sum_{\substack{1 \le j \le N \text{ and } 1 \le k \le N \\j \neq k}}
    \frac{ M(j,k)}{N(N-1)}
    \Bigg|.
    \label{eq:ddiag}
\end{equation}

The \textit{de-identification} metric is defined as  $\mathrm{DeID}= 1-{D_\text{diag}(M_\text{oa})}/{D_\text{diag}(M_\text{oo})}$ and it is expressed in percent.
$\mathrm{DeID}=100$\% means perfect de-identification, while $\mathrm{DeID}=0$\% means no de-identification.
\textit{Gain of voice distinctiveness} is defined as
$G_{\text{VD}} = 10\log_{10} \left( {D_\text{diag}(M_\text{aa})}/{D_\text{diag}(M_\text{oo})}\right)$,
where 0 means that the voice distinctiveness remains globally the same after anonymization, and a gain above or below 0 corresponds respectively to a global increase or a loss of voice distinctiveness.

\paragraph{\textbf{Word error rate (WER)}}

\begin{figure}[tp]
\centering\includegraphics[width=94mm]{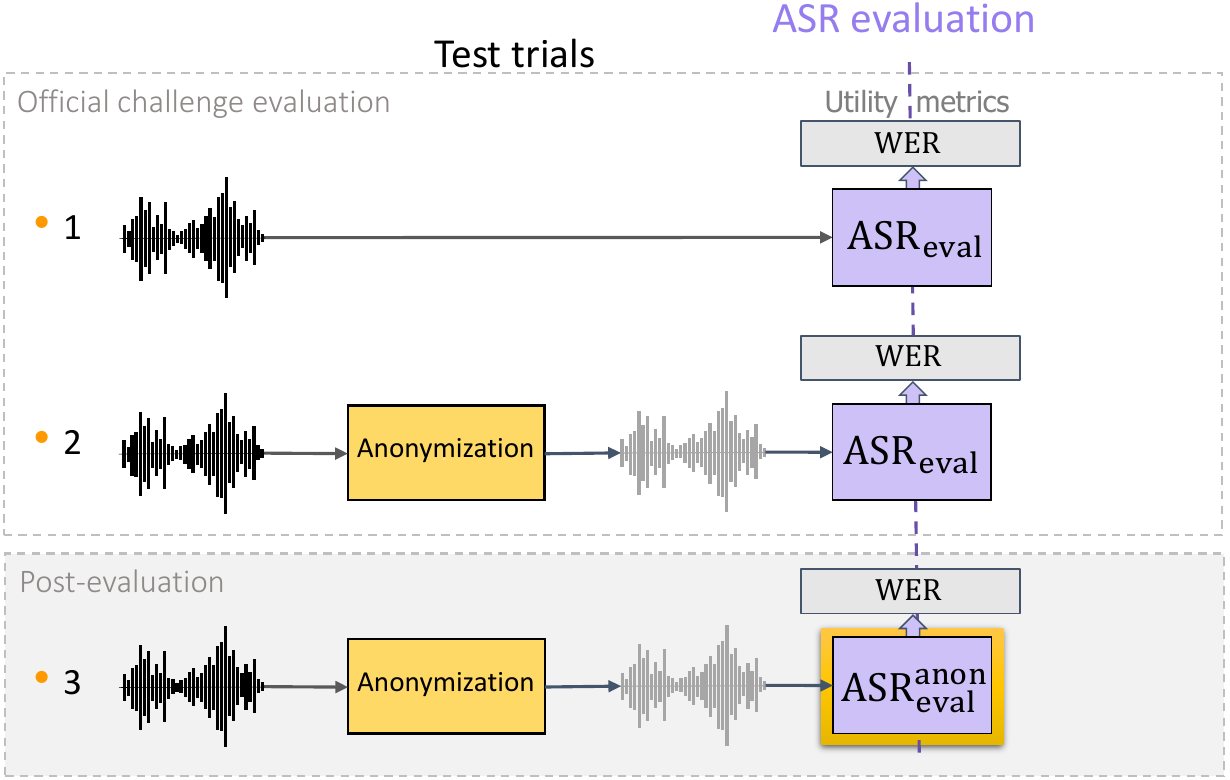}
\caption{ASR evaluation for the official challenge setup using $ASR_\text{eval}$ trained on original data is performed for two cases: (1) original  trial data and (2)  anonymized trial data.  ASR evaluation for the post-evaluation analysis is performed using $ASR_\text{eval}^\text{anon}$ trained on anonymized data for case (3) anonymized trial data. }
\label{fig:asr-eval}
\end{figure}

ASR performance is assessed using \textrm{$ASR_\text{eval}$} which is based on the adapted Kaldi recipe for LibriSpeech involving an acoustic model with  a factorized time delay neural network (TDNN-F)   architecture \cite{povey2018semi,peddinti2015time}, trained on the \textit{LibriSpeech-train-clean-360} dataset, and a trigram language model.
As shown in Figure~\ref{fig:asr-eval},  the (1) original and (2) anonymized trial data is decoded using the pretrained \textrm{$ASR_\text{eval}$} model and the WERs are calculated.
For the post-evaluation analysis, we also perform  decoding of anonymized trial data using the
$ASR_\text{eval}^\text{anon}$ model trained on anonymized data  (Figure~\ref{fig:asr-eval}, case 3).

\subsubsection{Subjective metrics}\label{sec:subj_eval_meth}

We consider two subjective privacy metrics (\emph{speaker verifiability} and \emph{speaker linkability}), and two subjective utility metrics (\emph{speech naturalness} and \emph{speech intelligibility}). \added[id=rev, comment=2.3]{The speaker verifiability and speech intelligibility metrics are subjective counterparts to the EER/$C_\text{llr}$/$C^\text{min}_\text{llr}$ and WER metrics, and aim to assess how human perception differs from objective evaluation. The speaker linkability metric provides a closer account of the way humans perceive voice characteristics and distinguish voices as belonging to certain speakers. Finally, the speech intelligibility metric is motivated by the requirement that the anonymized voices should sound natural, for which no established objective metric exists.}

\begin{figure}[t]
\centering
\includegraphics[width=0.95\linewidth]{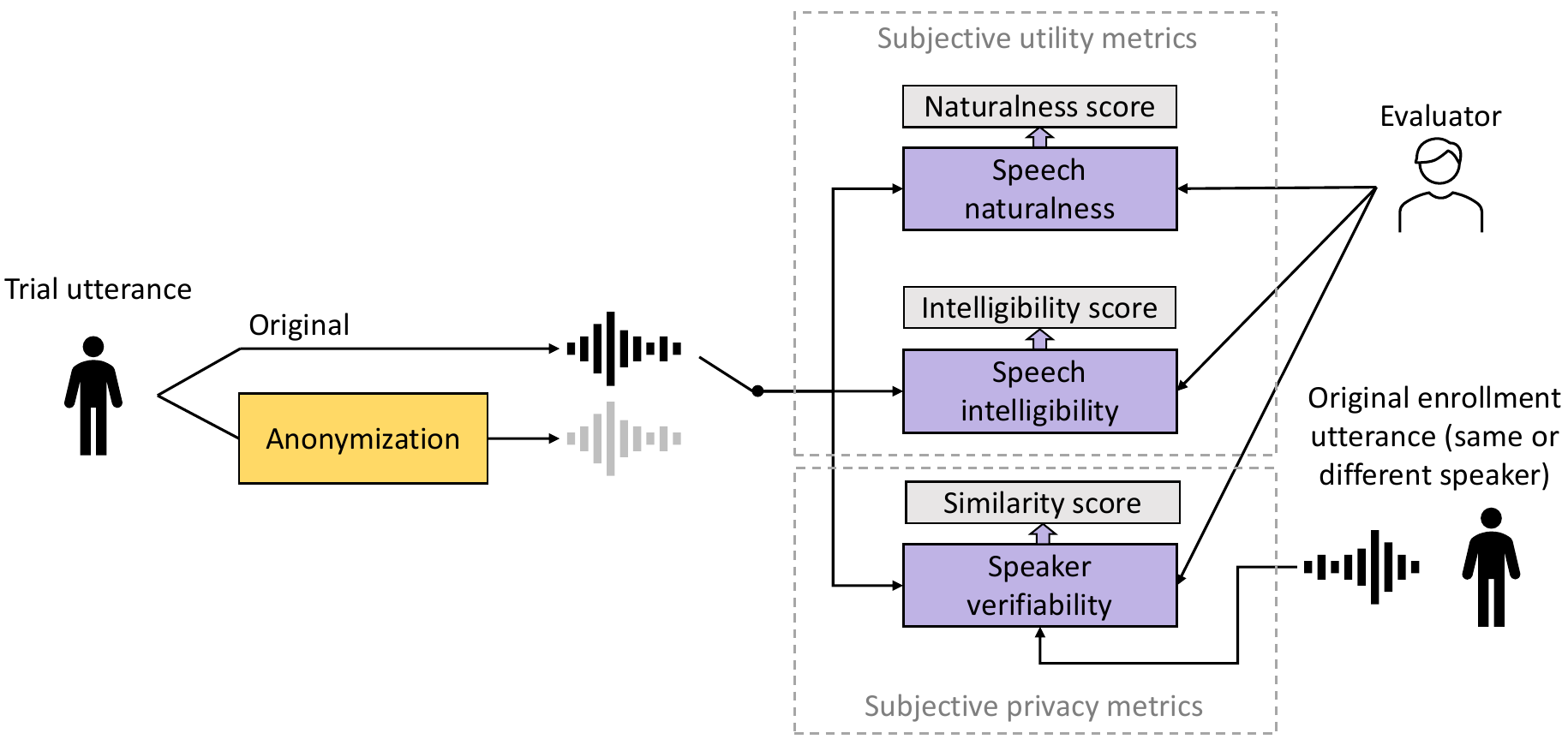}
\caption{Subjective evaluation test for speech naturalness, intelligibility, and speaker verifiability.}
\label{fig:sub-eval-design}
\end{figure}

\paragraph{\textbf{Subjective speech naturalness, intelligibility, and speaker verifiability}} 
These three metrics 
\added{were}
evaluated using the unified subjective evaluation test illustrated in Figure~\ref{fig:sub-eval-design}. \added{Each evaluator was asked to rate one
original or anonymized test set trial at a time.} 
For naturalness, 
the evaluator assigned a score from 1 (`totally unnatural') to 10 (`totally natural'). For intelligibility, the evaluator
\added{assigned}
a score from 1 (`totally unintelligible') to 10 (`totally intelligible'). For speaker verifiability, the evaluator 
\added{was}
required to listen to one original enrollment utterance from \added{the same or a different speaker} and rate the similarity between the trial and enrollment voices using a scale of 1 to 10, where 1 denotes `different speakers' and 10 denotes `the same speaker' with highest confidence. The evaluator was instructed to assign the scores through a role-playing game.\footnote{Details are given by 
\citet[Section~4.1]{Tomashenko2021CSlsupplementay}.}

Every evaluator was required to evaluate 36 trials in one session, following the procedures in Figure~\ref{fig:sub-eval-design}. He or she
\added{could}
also evaluate more than one session. The trials were randomly sampled from the speakers in the three test sets. The ratio of anonymized vs.\ original trials 
\added{was}
roughly 1:1. So 
\added{was}
the ratio of \added{enrollment-trial pairs from the same vs.\ different} speakers. Among the anonymized trials, the proportion of trials from each submitted anonymization system
\added{was}
also balanced. 
\deleted[id=rev, comment=1.10]{There are }
47 native English speakers participated in the evaluation and evaluated $16,200$ trials. The decomposed numbers of trials over the three test sets are listed in Table~\ref{tab:data-test}.

\begin{table}[ht]
  \caption{Number of trials for the subjective evaluation of speech naturalness, intelligibility, and speaker verifiability. 
  The anonymized trials are from 9 anonymization systems (2  baselines and 7 primary participants' systems).
   The number of speakers is 30 (15 male and 15 female) in each dataset, i.e., with respect to Table \ref{tab:data}, 
   2 male speakers were re-sampled 
   and 1 female speaker was discarded for \textit{LibriSpeech}.
  }
  \renewcommand{\tabcolsep}{0.13cm}
  \label{tab:data-test}
  \centering
    \resizebox{0.67\textwidth}{!}{
  \begin{tabular}{|c|r|r|r|r|}
    \hline
 \textbf{Test set} &   \textbf{Trials} &  \textbf{Female} & \textbf{Male} & \textbf{Total} \\ \hline \hline
\multirow{2}{*}{\shortstack{LibriSpeech \\ test-clean}}  
& Original &  \numprint{1330} &	 \numprint{1330}	&  \numprint{2660} \\
& Anonymized &  \numprint{1330} &	 \numprint{1330} &  \numprint{2660} \\  \hline
\multirow{2}{*}{\shortstack{VCTK-test \\ (common)}}  
& Original  &  \numprint{1380} &	 \numprint{1380}	&  \numprint{2760}\\ 
& Anonymized   &  \numprint{1380} &	 \numprint{1380}	&  \numprint{2760}\\ \hline
\multirow{2}{*}{\shortstack{VCTK-test \\ (different)}} 
& Original &  \numprint{1340} &	 \numprint{1340}	&  \numprint{2680}\\ 
& Anonymized  &  \numprint{1340} &	 \numprint{1340}	&  \numprint{2680}\\ \hline
  \end{tabular}}
\end{table}

\added{To reduce the perceptual bias of each evaluator, the scores were subject to normalized-rank normalization \cite{rosenberg2017bias}. The normalized scores are real-valued numbers in $[0, 1]$. The Mann-Whiteney-U test \cite{rosenberg2017bias} was used to assess statistical significance.}

  \paragraph{\textbf{Perception of speaker identity and speaker linkability}}
Evaluating the perception of speaker identity by humans is not simple. The subjective \added{verifiability and intelligibility scores} described above
\added[id=rev, comment=1.12]{closely mimic}
the corresponding objective metrics. Yet, the question whether they suffer from perceptual biases like the memorisation bias (\added{the evaluator recalls hearing the same voice previously}) or the well-known priming effect \added{(exposure to a stimulus inconsciently influences the response to a subsequent stimulus)} remains open. In order both to assess speaker linkability (i.e., the ability to cluster utterances into speakers) and to decrease as much as possible the influence of such biases, we designed a clustering-based perceptual experiment and the corresponding metrics. We developed a specific software tool for this purpose \cite{o2021anonymous}.\footnote{https://demo-lia.univ-avignon.fr/voiceprivacy/instructions} 

Due to the time-consuming nature of this experiment, only the two baseline anonymization systems were evaluated. 74 evaluators were recruited: 29 are native English speakers and the others are either bilingual or hold a high level of English. Each evaluator did only one session composed of three \added{panels}, resulting in a total of 222 \added{panels}. Each \added{panel} includes 16 utterances from 3 reference speakers (2 to 6 utterances each) and 1 distractor speaker (1 utterance only). Including a distractor helps to verify that the evaluators focus on speaker specificities and are not disturbed by other acoustic differences.
The anonymized distractor speaker was used to examine whether anonymization systems affect speaker discrimination performance, e.g., the evaluator either correctly identified the speaker as unique or incorrectly included it in a reference cluster.

For each \added{panel}, the evaluators 
\added{were}
asked to group the 16 utterances into 1 to 4 clusters according to subjective speaker voice similarity. In order to avoid perceptual biases as much as possible, during a given session, each speaker 
\added{was}
encountered in only 1 \added{panel}, and all speakers 
\added{were}
of the same gender. For the control \added{panel}, \replaced{original}{genuine} speech
\added{was}
used \added{for all utterances}; for the two other \added{panels}, half of the utterances 
\added[id=rev, comment=1.10]{were}
anonymized using the same anonymization system. 
 The data used in the speaker clustering task come from the  \textit{VCTK-test (common)} corpus. Unlike all other experiments, only the first 3~s of each utterance were used.
 \added[id=rev, comment=1.13]{The motivation for this length restriction was to provide evaluators with excerpts that were 
 short enough to not induce complex cognitive processes that involve complex syntactic, semantic, and pragmatic analysis. If the evaluators were provided longer excerpts, they could become distracted by attempting to complete and understand text narratives. In addition, limiting the duration of the excerpts reduces the risk of evaluator fatigue.}

As a primary metric, we use the macro-average \textit{F-measure} ($F_1$), a classical metric for such a task.
We also use a  secondary metric called \textit{clustering purity}.
 \textit{Clustering purity} associates each cluster with a unique ground truth speaker and focuses only on precision, 
while $F_1$ allows two clusters to correspond to the same ground truth speaker and is the harmonic mean of precision and recall.
Clustering \textit{purity}  is defined as
\begin{equation}\label{eq:purity}
purity(C) = \max_{s \in S}\frac{1}{N}\sum_{c \in C} |{c\cap s_c}|,
\end{equation}
where $C$ is the set of estimated clusters, $c$ is an individual cluster in $C$, $S$ is the set of all possible
combinations of unique speakers assigned to each cluster,  $s_c$ is the speaker label assigned to cluster $c$ in combination $s$, and $N$ is the number of utterances in the \added{panel}. 
\added{In addition, we consider a \textit{clustering change} (CC) metric, that is the number of times an evaluator (re-)assigns an utterance to a cluster.}

\section{Anonymization systems}\label{sec:system_description}

We now describe the two baseline systems provided by the challenge organizers as well as those prepared by challenge participants.

\subsection{Baseline systems}\label{sec:baseline}

Two different anonymization systems were provided as challenge baselines\footnote{  \url{https://github.com/Voice-Privacy-Challenge/Voice-Privacy-Challenge-2020}} \added[id=rev, comment=2.3]{to help the participants tackle this relatively new task and explore a wide range of solutions. The first baseline offers more flexibility in the choice of the pseudo-speaker and provides state-of-the-art objective privacy and utility, but it requires significant development efforts and big computational resources. In contrast, the second baseline is simpler and provides good subjective speech naturalness and intelligibility, but it results in weaker privacy preservation.}

\begin{figure}[h]
\centering\includegraphics[width=100mm]{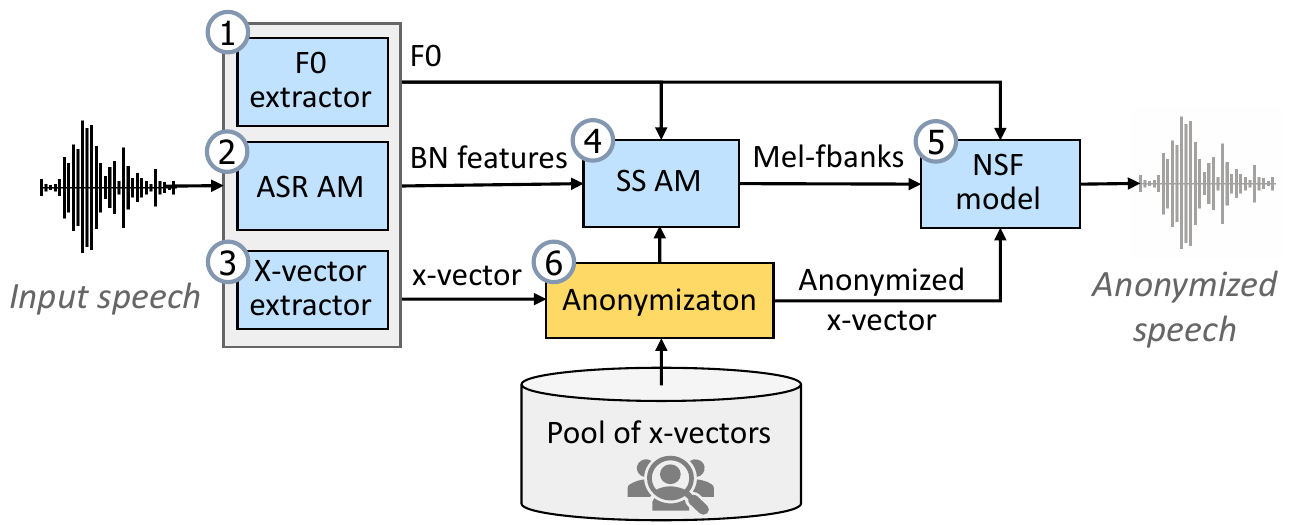}
\caption{Primary baseline  anonymization system (\textbf{B1}). 
}
\label{fig:baseline1}
\end{figure}

\label{sec:baseline-1}
The \emph{primary} baseline, denoted \textbf{B1}, is shown in Figure~\ref{fig:baseline1}.  It is inspired from \citet{fang2019speaker}
and performs 
anonymization using x-vectors \cite{snyder2018x} and neural speech synthesis.
It comprises three steps: (1)~x-vector, pitch~(F0) and bottleneck~(BN) feature extraction; (2)~x-vector anonymization; (3)~speech synthesis~(SS) using the anonymized x-vector and the original F0 and BN features.
Step~(1) 
encodes the spoken content by 256-dimensional BN features extracted using a TDNN-F ASR AM trained on the \textit{LibriSpeech train-clean-100} and \textit{train-other-500} datasets and speaker information by a 512-dimensional x-vector extracted using a TDNN trained on the \textit{VoxCeleb-1,2} dataset. Both extractors are implemented with the Kaldi toolkit.
Step~(2) computes an anonymized x-vector for every original x-vector. It is generated by averaging a set of $N^*$ x-vectors selected at random from a larger set of $N$  x-vectors, itself composed of the $N$ farthest x-vectors in the \textit{LibriTTS train-other-500} dataset, according to PLDA distance.\footnote{In the baseline, we use $N=200$ and $N^*=100$.}
Step~(3) uses a SS AM to generate Mel-filterbank features from the anonymized x-vector and the original F0 and BN features, and a neural source-filter (NSF) waveform model \cite{wang2019neural} to synthesize a speech signal from the anonymized x-vector and the F0 and Mel-filterbank features. The SS AM and NSF models are both trained on the \textit{LibriTTS train-clean-100} dataset. 
\added[id=rev,comment=2.4]{With respect to the work by \citet{fang2019speaker}, the differences in baseline B1 include using PLDA distance instead of cosine distance and using a different x-vector selection strategy. Also, the model architectures for each step and the training datasets differ.}
Full details are provided by \citet{tomashenkovoiceprivacy}. \citet{srivastava2020baseline} evaluate these design choices against other possible choices.

\label{sec:baseline-2}

\begin{figure}[htp]
    \centering
    \includegraphics[width=105mm]{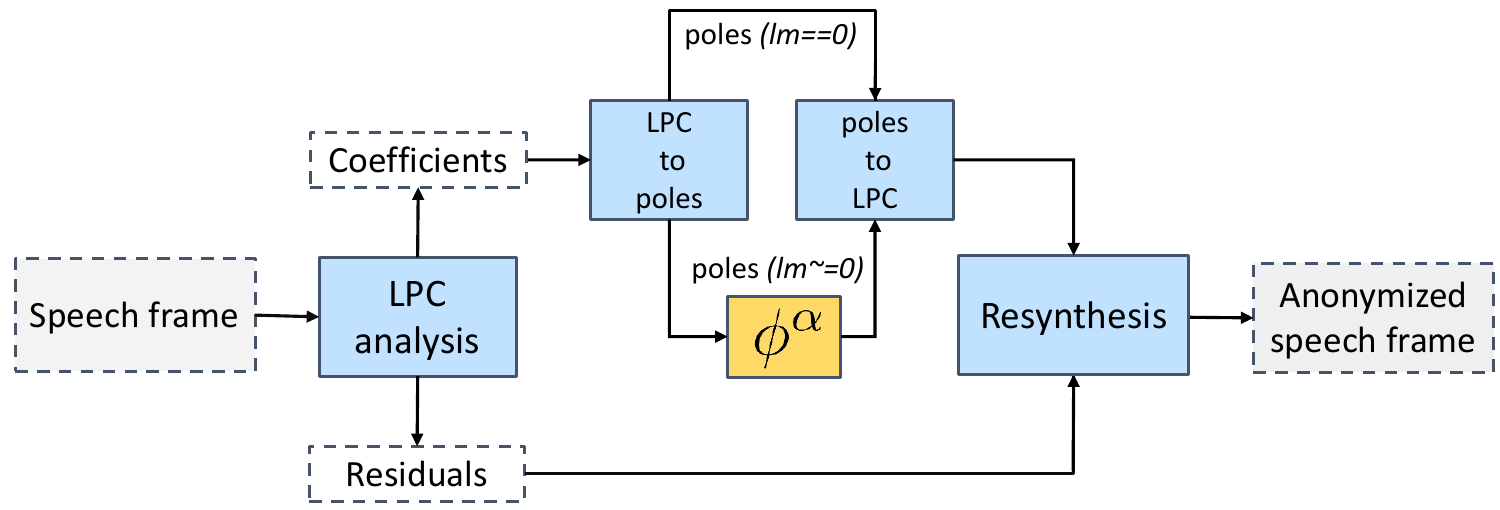}
    \caption{Secondary baseline anonymization system (\textbf{B2}). 
    }
    \label{fig:lpc_processing}
\end{figure}

In contrast to the primary baseline, the \emph{secondary} baseline, denoted \textbf{B2},
does not require any training data and is based upon traditional signal processing techniques \cite{patino2020speaker}. It employs the McAdams' coefficient \cite{mcadams1984spectral} to achieve anonymization by shifting the pole positions derived from the linear predictive coding (LPC) analysis of speech signals. 
The process is depicted in Figure~\ref{fig:lpc_processing}.  It starts with the application of frame-by-frame LPC source-filter analysis to derive LPC coefficients and residuals. The residuals are set aside for later resynthesis, \added[id=rev, comment=2.5]{whereas LPC coefficients are converted into pole positions by polynomial root-finding.}
 The McAdams' transformation is then applied to the angles of the poles (with respect to the origin in the z-plane), each one of which corresponds to a peak in the spectrum (resembling formant positions). While real-valued poles are left unmodified, the angles $\phi$ of the poles with a non-zero imaginary part (with values between 0 and $\pi$ radians) are raised to the power of the McAdams' coefficient $\alpha$ so that the transformed pole has new, shifted angle $\phi^\alpha$.
The value of $\alpha$ implies a contraction or expansion of the pole positions around $\phi=1$.
For a sampling rate of 16~kHz, i.e.\ for  data used in the challenge, $\phi=1$ corresponds to approximately 2.5~kHz which is the approximate mean formant position \cite{ghorshi2008cross}.
Corresponding complex conjugate poles are similarly shifted in the opposite direction  
and the new set of poles, including original real-valued poles, 
are then converted back to LPC coefficients. Finally, LPC coefficients and residuals are used to resynthesise a new speech frame in the time domain.
\added[id=rev,comment=2.4]{This technique shares some similarities with the frequency warping based methods previously explored by \citet{qian2018hidebehind} and \citet{srivastava2019privacy} except that, for the sake of simplicity, it modifies only the spectral envelope (not the pitch).}
Full details are provided by \citet{patino2020speaker}.

\subsection{Submitted systems}\label{sec:sys_participants}

The VoicePrivacy Challenge attracted 45 participants from both academic and industrial organizations and 13 countries,  representing 25 teams. 
Among the 5 allowed submissions by each team, participants were required to designate one as their primary system with any others being designated as contrastive systems.  With full descriptions available elsewhere, we provide only brief descriptions of the 16 successful, eligible submissions, a summary of which is provided in~Table~\ref{tab:submissions} which shows system identifiers (referred to below) in column~3.
Most systems submitted to the VoicePrivacy 2020 challenge were inspired by the primary baseline 
(see Section~\ref{sec:submissions_from_b1}).  One submission is based upon the secondary baseline (see Section~\ref{sec:submissions_from_b2}) whereas two others are not related to either (see Section~\ref{sec:submissions_other}).\footnote{There is also one non-challenge entry work 
related to the challenge \cite{huang2020analysis}. This team worked on the development of stronger attack models for ASV evaluation.}

\newlength{\dpcircle}
\newlength{\rcircle}
\newlength{\dcircle}
\newcommand{\docircle}[4]{%
  \setlength{\dpcircle}{\dp\strutbox}%
  \setlength{\dcircle}{\dpcircle}%
  \addtolength{\dcircle}{\ht\strutbox}%
  \setlength{\rcircle}{0.5\dcircle}%
  \setlength{\unitlength}{1sp}%
  \begin{picture}(\number\dcircle,0)
    \color{#1}
    \put(\number\rcircle,\number\dpcircle){\circle*{\number\dcircle}}
    \color{#2}
    \put(\number\rcircle,\number\dpcircle){\circle{\number\dcircle}}
    \put(\number\rcircle,0){\makebox[0pt]{\textcolor{#3}{#4}}}
  \end{picture}%
}

\newcolumntype{L}{m{9.8cm}}
\newcommand\encircle[1]{%
  \tikz[baseline=(X.base)] 
    \node (X) [draw, shape=circle, inner sep=0] {\strut #1};}

\begin{table}[H]
  \caption{Teams, organizations, and submitted systems. The submission identifier (ID) for each system in the last column comprises: $<$team id: first letter of the team name$><$submission deadline\protect\footnotemark: 1 or 2$><$c, if the system is contrastive$><$index of the contrastive system$>$. The symbol $\color{blue}\star$ in the first column indicates that the team submitted the anonymized training data for post-evaluation analysis. The colors \docircle{babyblue}{black}{black}{1} and \docircle{bubblegum}{black}{black}{2} indicate systems that were developed from \textbf{B1} or \textbf{B2}, respectively, while \docircle{gainsboro}{black}{black}{} indicates other systems.}\label{tab:submissions}.
\renewcommand{\tabcolsep}{0.095cm}
  \centering
  \resizebox{1.0\textwidth}{!}
{
  \begin{tabular}{|c|L|l|} 
\hline
\textbf{Team (Reference) } & \textbf{Organization(s)} & \textbf{Sys.}  \\ \hline \hline
\makecell[c]{AIS-lab JAIST \\ \cite{Mawalim2020} \\  \large \docircle{babyblue}{black}{black}{A} $\color{white}\star$ }	& \makecell[L]{$\bullet$Japan Advanced Institute of Science and Technology,  Japan \\ $\bullet$NECTEC, National Science and Technology Development Agency, Thailand }	& \makecell[l]{A1 \\ A2} \\ \hline
\makecell[c]{ DA-IICT Speech Group \\  \cite{gupta2020design} \\  {\large\docircle{bubblegum}{black}{black}{D}} $\color{white}\star$ \\  }	& \makecell[L]{$\bullet$Dhirubhai Ambani Institute of Information and Communication Technology, India}	& \makecell[l]{D1}  \\ \hline
\makecell[c]{Idiap-NKI \\ \cite{dubagunta2020adjustable} \\  {\large\docircle{gainsboro}{black}{black}{I}}} $\color{white}\star$& \makecell[L]{$\bullet$Idiap Research Institute, Martigny, Switzerland \\
 $\bullet$École Polytechnique Fédérale de Lausanne (EPFL), Switzerland \\
 $\bullet$Netherlands Cancer Institute (NKI), Amsterdam, Netherlands \\}	& \makecell[l]{I1}  \\ \hline
\makecell[c]{Kyoto Team \\ \cite{kyoto2020} \\  \large\docircle{gainsboro}{black}{black}{K} $\large\color{blue}\star$}	& \makecell[L]{ $\bullet$Kyoto University, Kyoto, Japan \\
$\bullet$National Institute of Information and Communications Technology, Kyoto, Japan}	& \makecell[l]{K2} \\ \hline
\makecell[c]{MultiSpeech \\	\cite{champion2020speaker} \\ \\ \large\docircle{babyblue}{black}{black}{M} $\color{blue}\star$}  & \makecell[L]{$\bullet$Université de Lorraine, CNRS, Inria, LORIA, Nancy, France \\  $\bullet$Le Mans Université, LIUM, France}	& \makecell[l]{M1 \\ M1c1 \\ M1c2  \\ M1c3 \\ M1c4 } \\ \hline
\makecell[c]{Oxford System Security Lab \\ \cite{turner2020speaker}  \\ \large\docircle{babyblue}{black}{black}{O} $\color{blue}\star$  \\} 	& \makecell[L]{ $\bullet$University of Oxford, UK}	& \makecell[l]{O1 \\ O1c1 }  \\ \hline
\makecell[c]{Sigma Technologies SLU \\ \cite{espinoza2020speaker} \\  \large\docircle{babyblue}{black}{black}{S} $\color{blue}\star$} 	& \makecell[L]{$\bullet$Sigma Technologies S.L.U., Madrid, Spain \\
$\bullet$Universidad Politecnica de Madrid, Spain}	& \makecell[l]{S1 \\ S1c1 \\ S2 \\ S2c1}  \\ \hline
\makecell[c]{\textcolor{gray}{PingAn} \\ \textcolor{gray}{\cite{huang2020analysis}}}	& \makecell[L]{ \textcolor{gray}{\color{lightgray}{$\color{lightgray}{\bullet}$}\textcolor{gray}{PAII Inc., Palo Alto, CA, USA}} }	& \textcolor{lightgray}{-}  \\  \hline
\end{tabular}}
\end{table}
\normalsize
\footnotetext{deadline-1: 8th May 2020; deadline-2: 16th June 2020.}

\subsubsection{Submissions derived from Baseline-1}\label{sec:submissions_from_b1}

Teams \textbf{A}, \textbf{M},  \textbf{O} and \textbf{S} (see identifiers in column~3 of Table~\ref{tab:submissions} and column~1 of Table~\ref{tab:submissions-b1}) submitted systems derived from the primary baseline.  
Table~\ref{tab:submissions-b1} provides an overview of the modifications made by each team to the baseline modules shown in Figure~\ref{fig:baseline1}.
None of the teams modified the x-vector extraction module (\#3 in Table~\ref{tab:submissions-b1}), whereas 
two systems modified the x-vector anonymization module (\#6).
Details of specific modifications are described in the following.  We focus first on differences made to specific modules, then on specific system attributes.

\newcolumntype{i}{m{7.8cm}}

\begin{table}[H]
  \caption{Summary of the challenge submissions derived from \textbf{B1}. \colorbox{babyblue}{$\checkmark$} and \colorbox{babyblue}{blue color}  indicate the components and speaker pool data that were modified 
  w.r.t.  \textbf{B1}.
  }\label{tab:submissions-b1}.
\renewcommand{\tabcolsep}{0.0695cm}
  \centering
  \resizebox{1\textwidth}{!}{
  \begin{tabular}{|l|i|l|l|l|l|l|l|l|} 
\hline
\small
\multirow{ 2}{*}{\makecell{\textbf{Sys.}}} & \multirow{ 2}{*}{\textbf{Description of modifications}} & \textbf{1} & \textbf{2} &  \textbf{3} &  \textbf{4} & \textbf{5} & \textbf{6} 
&\multirow{2}{*}{\makecell{\textbf{Data for} \\ \textbf{       speaker pool}}}  \\ \cline{3-8}
 &  & \small\rotatebox{90}{F0 } & \small\rotatebox{90}{ASR} &  \small\rotatebox{90}{X-vect. } &  \small\rotatebox{90}{SS} & \small\rotatebox{90}{NSF }& \small\rotatebox{90}{Anon. } 
&  \\ \hline \hline
A2 & \small\small{using singular value modification}   & & &  \textcolor{white}{$\checkmark$} & & & \cellcolor{babyblue} $\checkmark$ & \cellcolor{babyblue}  \makecell{
LibriTTS: \\train-other-500 \tiny{~}} \\ \cline{1-8}
A1 &  \small{\small{\textcolor{gray}{different F0 extractor\protect\footnotemark}; x-vector anonymization using variability-driven ensemble regression modeling}} & \cellcolor{babyblue} $\checkmark$ & & & & &\cellcolor{babyblue}  $\checkmark$ & \cellcolor{babyblue} \makecell{LibriTTS: \\  train-clean-100\small{~} } \\  \hline
M1 &  \small{\small{End-to-end ASR AM}}& & \cellcolor{babyblue} $\checkmark$ & & \cellcolor{babyblue} $\checkmark$ & \cellcolor{babyblue} $\checkmark$ & & \\ \hline
M1c1 & \small{\small{End-to-end ASR AM; semi-adversarial training  to learn linguistic features while masking speaker information}} & & \cellcolor{babyblue} $\checkmark$ & & \cellcolor{babyblue} $\checkmark$ & \cellcolor{babyblue} $\checkmark$ & & \\ \hline
M1c2 & \small{\small{copy-synthesis (original x-vectors)}} & & &  & & & \cellcolor{babyblue} $\checkmark$ & \\ \hline
M1c3 & \small{\small{x-vectors provided to SS AM are anonymized, x-vectors provided to NSF are original }} & & &  & &  & \cellcolor{babyblue} $\checkmark$ & \\ \hline
M1c4 & \small{\small{x-vectors provided to SS AM are original, x-vectors provided to NSF are anonymized}} & & &  & & & \cellcolor{babyblue} $\checkmark$ & \\ \hline
O1   & \small\small{keeping original distribution of cosine distances between speaker x-vectors; GMM for sampling speaker vectors in a PCA-reduced space followed by projection to the original x-vector dimension}& & & & & & \cellcolor{babyblue} $\checkmark$ & \cellcolor{babyblue}  {\makecell[c]{LibriTTS:\\ train-other-500\tiny{~}  }} \\ \cline {1-8}
O1c1 &  \small\small{\small\textbf{O1} with forced dissimilarity between original and generated x-vectors}& & & & & & \cellcolor{babyblue} $ \cellcolor{babyblue} \checkmark$ &  \makecell[c]{~VoxCeleb - 1,2\large{~}} \cellcolor{babyblue} \\ \hline
S1   & \small{\textbf{S1c1} applied on the top of the \textbf{B1} x-vector anonymization}& & & & & & \cellcolor{babyblue} $\checkmark$ & \\ \hline
S1c1 & \small{domain-adversarial training; autoencoders: using gender, accent, speaker id outputs corresponding to adversarial branches in ANN for x-vector reconstruction} & & & & & &  \cellcolor{babyblue} $\checkmark$ & \\ \hline
S2   & \small{\textbf{S2c1} applied on the top of the \textbf{B1} x-vector anonymization} & & & & & & \cellcolor{babyblue} $\checkmark$ & \\ \hline
S2c1   & \small{\textbf{S1c1} with parameter optimization} & & & & & &  \cellcolor{babyblue} $\checkmark$ & \\ \hline
\end{tabular}}
\end{table}
\normalsize
\footnotetext{Different F0 extractors were used in experiments, but the baseline F0 in the final \textbf{A1}.}

\paragraph{\textbf{F0:}} Only team \textbf{A} \cite{Mawalim2020} modified the pitch extractor. 
They replaced the baseline F0 extractor with WORLD 
\cite{morise2016world} and by SPTK\footnote{Speech Signal Processing Toolkit (SPTK): \url{http://sp-tk.sourceforge.net/}} 
alternatives. 
While no significant impact upon ASR performance was observed, 
SPTK F0 estimation was found to have some impact, albeit inconsistent, upon the ASV EER.
Consequently, the final system used the baseline F0 extractor.
Post-evaluation work conducted by \citet{champion2020astudyf0} showed improved anonymization performance when
F0 statistics of the original speaker are replaced with those of a pseudo-speaker, without significant impact upon the ASR performance.

\paragraph{\textbf{ASR AM, speech synthesis AM and NSF model:}} 
Instead of the baseline hybrid TDNN-F ASR acoustic model,
systems \textbf{M1} and \textbf{M1c1} \cite{champion2020speaker} used an end-to-end model with a hybrid connectionist temporal classification (CTC) and attention architecture \cite{watanabe2017hybrid} for BN feature extraction.
The SS AM and NSF models were then re-trained using the new BN features.
In addition, the \textbf{M1c1} contrastive system relied on semi-adversarial training of the ASR AM to learn linguistic features while masking speaker information.

\paragraph{\textbf{X-vector anonymization:}} 
All teams explored different approaches to x-vector anonymization.  They are described in the following:
\paragraph{\textbf{$\circ$A2}}  
\emph{Singular value modification} \cite{Mawalim2020}. The singular value decomposition (SVD) of the matrix 
constructed from the utterance-level speaker x-vectors was  used for anonymization. The target x-vector was obtained from the least similar centroid using x-vector clustering.
Anonymization was performed through modification of the matrix singular values. A singular value threshold parameter determines the dimensionality reduction used in the modification and determines the percentage of the kept non-zero singular values.

\paragraph{\textbf{$\circ$A1}}
\emph{Variability-driven decomposition with regression models} \cite{Mawalim2020}. The  speaker x-vector was decomposed into high- and low-variability components which were separately modified using two different regression models.
It was argued that speaker-specific information is mostly contained in the low-variability component, which is hence the component upon which the anonymisation must focus.
\paragraph{\textbf{$\circ$O1}} \emph{Distribution-preserving x-vector generation} \cite{turner2020speaker}. 
Baseline \textbf{B1} performs anonymization through x-vector averaging.  As a result, the anonymized voices are less diverse than the original voices and the resulting differences in the distribution of original vs.\ anonymized x-vectors leaves the anonymization system vulnerable to inversion. \citet{turner2020speaker} investigated the use of GMMs to sample x-vectors in a PCA-reduced space in a way that retains the original distribution of cosine distances between speaker x-vectors, thereby improving robustness to inversion.

\paragraph{\textbf{$\circ$O1c1}} \emph{Forced dissimilarity between original and anonymized x-vectors} \cite{turner2020speaker}.
In a slight variation to the \textbf{O1} system, the \textbf{O1c1}
contrastive system generates a new x-vector in the case when the original and anonymized x-vectors are not sufficiently dissimilar.

\paragraph{\textbf{$\circ$S1c1 \& S2c1}} \emph{Domain-adversarial training} \cite{espinoza2020speaker}.
Domain adversarial training was used to generate x-vectors with separate gender, accent, and speaker adversarial branches in an autoencoder adversarial network.
For system \textbf{S2c1}, the parameters of the adversarial branches were tuned to optimise the trade-off between the autoencoder and the adversarial objectives.

\paragraph{$\circ$\textbf{S1 \& S2}} \emph{Domain-adversarial training on top of B1} \cite{espinoza2020speaker}.
The primary systems 
\textbf{S1} and \textbf{S2} are based upon the application of the 
contrastive systems \textbf{S1c1} and \textbf{S2c1} to the anonymized x-vectors generated by baseline \textbf{B1}.

\paragraph{$\circ$\textbf{M1c2}} \emph{Copy-synthesis} \cite{champion2020speaker}.
This contrastive system 
is essentially the \textbf{B1} baseline, but without \emph{explicit} x-vector anonymization,  It provides some insights into the added benefit of the latter, beyond simple copy-synthesis.

\paragraph{$\circ$\textbf{M1c3}} \emph{Original x-vectors for 
NSF}.
Another contrastive system for which
the NSF model receives original x-vectors while the SS AM receives anonymized x-vectors.

\paragraph{$\circ$\textbf{M1c4}} \emph{Original x-vectors for SS AM}.
A variation on the above contrastive systems whereby 
the SS AM receives original x-vectors but the
NSF model receives anonymised x-vectors.

\paragraph{$\circ$\textbf{\textbf{A} and \textbf{O}}} \emph{Speaker pool augmentation}.
In addition to their respective modifications made to x-vector anonymization, some teams also investigated the augmentation of the x-vector pool using additional datasets, namely 
\textit{LibriTTS-train-clean-100} (team \textbf{A}) and \textit{VoxCeleb-1,2} (team \textbf{O}).

\subsubsection{Submission derived from Baseline-2}\label{sec:submissions_from_b2}

\paragraph{$\circ$\textbf{D1}} \emph{Modifications of the pole radius} \cite{gupta2020design}.
Team \textbf{D} investigated modifications of the pole radius (distance from the origin) in addition to the shift in phase operated by baseline \textbf{B2}.
This approach further
distorts the spectral envelope.
Pole radii were reduced to 
0.975 of the original values whereas
the McAdams' coefficient was set to 0.8 as in baseline \textbf{B2}.

\subsubsection{Other submissions}\label{sec:submissions_other}

\paragraph{$\circ$\textbf{K2}} \emph{Anonymization using x-vectors, SS models and a voice-indistinguishability metric} \cite{kyoto2020}.
Similar to the primary baseline~\textbf{B1}, system \textbf{K2} is also based on x-vector anonymization, but the anonymization process and SS models (and corresponding input features) are quite different from those of baseline \textbf{B1}. Other differences include using the test dataset for creating the speaker pool.
The speech synthesis framework uses two modules:
(1)~an end-to-end AM implemented with ESPnet\footnote{\url{ https://github.com/espnet/espnet/tree/master/egs/librispeech/tts1}}
which 
produces a Mel-spectrogram from filterbank features and speaker x-vectors; 
(2)~a waveform vocoder based on the Griffin-Lim algorithm \cite{griffin1984signal}
which produces a speech waveform from the Mel-spectrogram after conversion to a
linear scale spectrogram.
A voice indistinguishability metric \cite{han2020voice} inspired by differential
privacy concepts \cite{dwork2009differential} was applied during x-vector perturbation to select target speaker x-vectors.

\paragraph{$\circ$\textbf{I1}} \emph{Modifications to formants, F0 and speaking rate} \cite{dubagunta2020adjustable}. 
The \textbf{I1} system is based upon a signal-processing technique inspired from \citet{robvanson2020adjustable}.
The playback speed was adjusted
to linearly shift formant frequencies.
Individual formants were then shifted to specific target values chosen 
from a set of randomly chosen speakers in
the \textit{LibriSpeech-train-other-500} dataset.
The F0 and the speaking rate were
also adjusted using a pitch-synchronous overlap-and-add method \cite{moulines1990pitch}. 
Additional processing includes exchanging the F4 and F5 bands using a Hann filter method and adding modulated pink noise to the
speaker F6--F9 bands for formant  masking.

\section{Results}\label{sec:results}

In this section we report the evaluation results for the systems described in Section~\ref{sec:system_description}. \added[id=rev, comment=1.8]{The results obtained as part of the challenge and those obtained as part of the post-evaluation analysis are both presented without distinction.}

\subsection{Objective evaluation results}\label{sec:obj_results}

We first present and discuss the objective evaluation results.

\subsubsection{Privacy: objective speaker verifiability}\label{sec:obj_results_off}

\begin{figure}[H]
\centering\includegraphics[width=\textwidth]{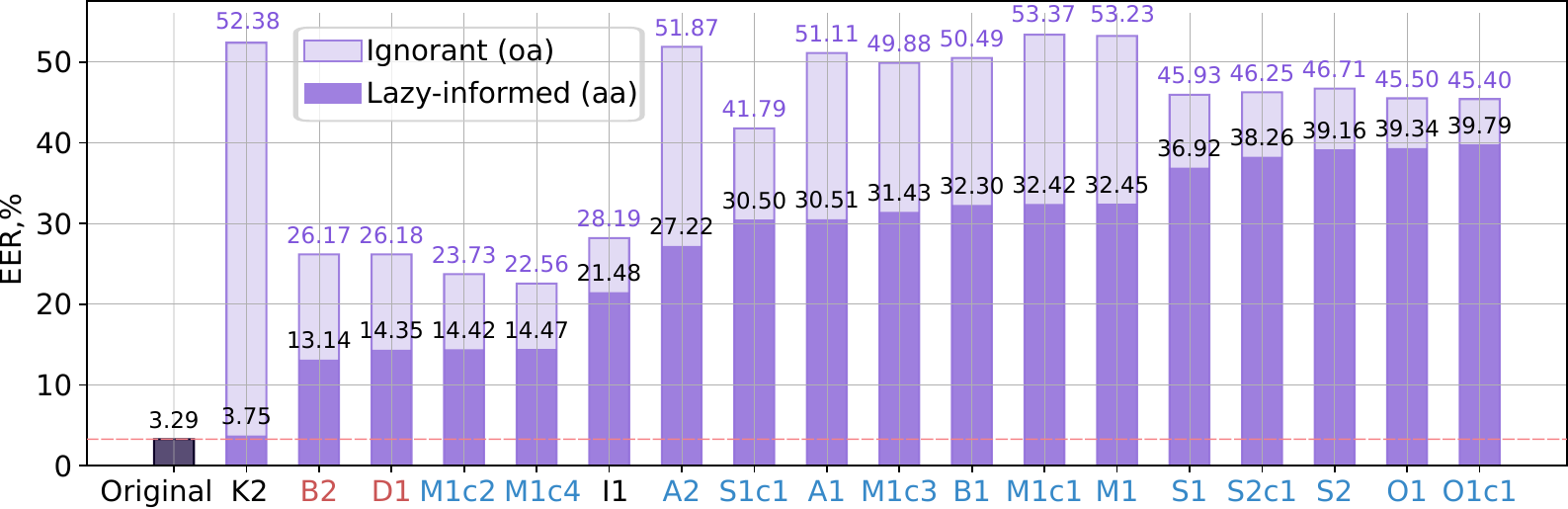}
\caption{Average EER over all test datasets for all anonymization systems and for original data, against ignorant (\textbf{oa}) or lazy-informed (\textbf{aa}) attackers. \textcolor{frenchblue}{Blue} and \textcolor{chestnut}{red} colors in the system IDs indicate systems developed from \textcolor{frenchblue}{\textbf{B1}} or  \textcolor{chestnut}{\textbf{B2}}, respectively. Higher EER corresponds to better privacy.}.
\label{fig:eer-mean}
\end{figure}

Speaker verifiability results are shown in Figure~\ref{fig:eer-mean} in terms of EER averaged across all  test datasets for the \textit{ignorant} (\textbf{oa}) and \textit{lazy-informed} (\textbf{aa}) attack models described in Section~\ref{subsec:attack_model}.
Without anonymization, the EER is $3.29\%$. Anonymization is
expected to increase the EER.

When only trial data is anonymized 
(\textbf{oa} condition, light bars in Figure~\ref{fig:eer-mean}), the EER increases for all anonymization systems: from $22.56\%$ for \textbf{M1c4} to $53.37\%$ for \textbf{M1c1}.
Better anonymization is achieved by using x-vector based anonymization systems
 (\textbf{K2, A*, S*, M*, B1, O*}) than signal processing based ones  (\textbf{B2, D1, I1}).
Systems \textbf{M1c2} and \textbf{M1c4} perform worst as expected, because they provide non-anonymized x-vectors to the speech synthesis AM, but they still result in an increased EER compared to original speech due to the acoustic mismatch between original and synthesised speech. 
Systems \textbf{K2, A*, M1c1, M1, B1} all produce EERs above $50\%$, indicating that the anonymization requirement against ignorant attackers is fully met.

Anonymization of both enrollment and trial utterances (\textbf{aa} condition, darker bars in Figure~\ref{fig:eer-mean}) results in universally lower EERs for all systems. 
While system rankings are different for each attack model, the general trend is consistent: \textbf{B1} based systems outperform others.  
Some results are of particular interest.
The EER of $3.75\%$ for system \textbf{K2} is only marginally above the original EER of $3.29\%$, despite it being the 3rd best performing system for the \textbf{oa} condition.
Even the best \textbf{O1c1} system achieves an EER of only $37.79\%$, which is far away from the $50\%$ which indicates successful anonymization.
These results highlight the importance of designing anonymization systems under the assumption of a strong attack model. Without it, results may provide a false sense of protection.

Overall, taking confidence intervals (not shown in the figure) into account, baseline \textbf{B1} is outperformed by systems \textbf{A1}, \textbf{A2}, \textbf{M1},  \textbf{M1c1}, and \textbf{K2} in the \textbf{oa} condition and by systems \textbf{S2}, \textbf{S2c1}, \textbf{O1}, and \textbf{O1c1} in the \textbf{aa} condition.
These two sets of systems do not intersect and no single system works best in both conditions.
This highlights the difficulty of designing and optimising an anonymization system that works well under different attack scenarios.
The results for system \textbf{K2} are also of note. This system
 achieves a very high anonymization performance in the \textbf{oa} condition due to the fact that anonymized utterances are acoustically very different from the original ones.
At the same time, it achieves a very poor performance in the \textbf{aa} condition since, instead of generating anonymized x-vectors from a
 dataset with many speakers (relative to the evaluation dataset), it generates them from the evaluation dataset itself.
This results in distinct confusions between some speakers, however the number of such confusions is very low, especially for some test sets (see, for example, the speaker similarity matrix $M_\text{aa}$ for female speakers on the \textit{LibriSpeech-test} set in Figure~\ref{fig:mat-libri-f-k2}).

\begin{figure}[h!]
\begin{center}
\begin{subfigure}{0.47\textwidth}
\center{\includegraphics[width=1\textwidth]{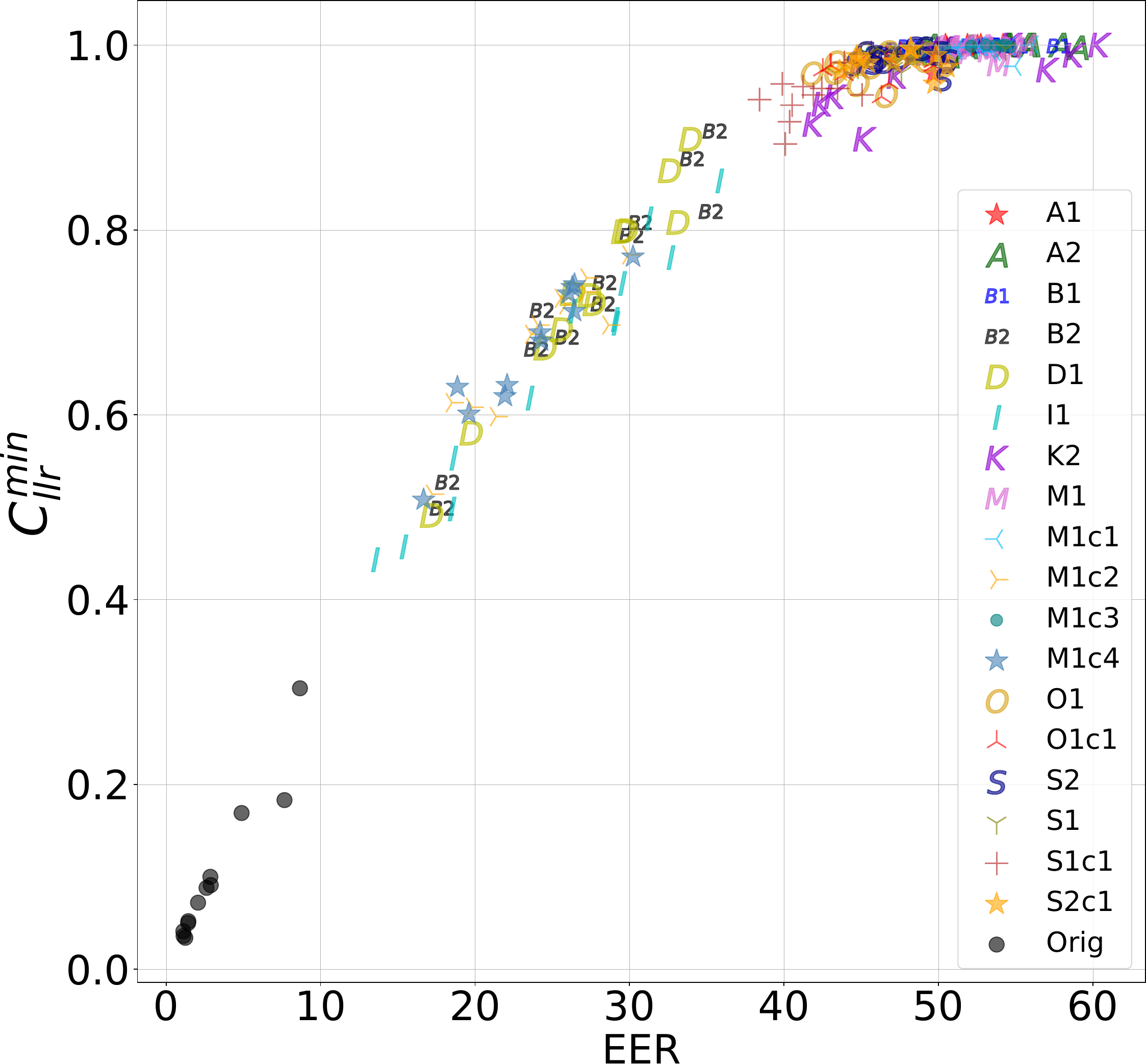}}
\caption{Ignorant attacker (oa)}
\label{fig:eer-vs-cllr-min-oa}
\end{subfigure}
~~~
\begin{subfigure}{0.47\textwidth}
\center{\includegraphics[width=1\textwidth]{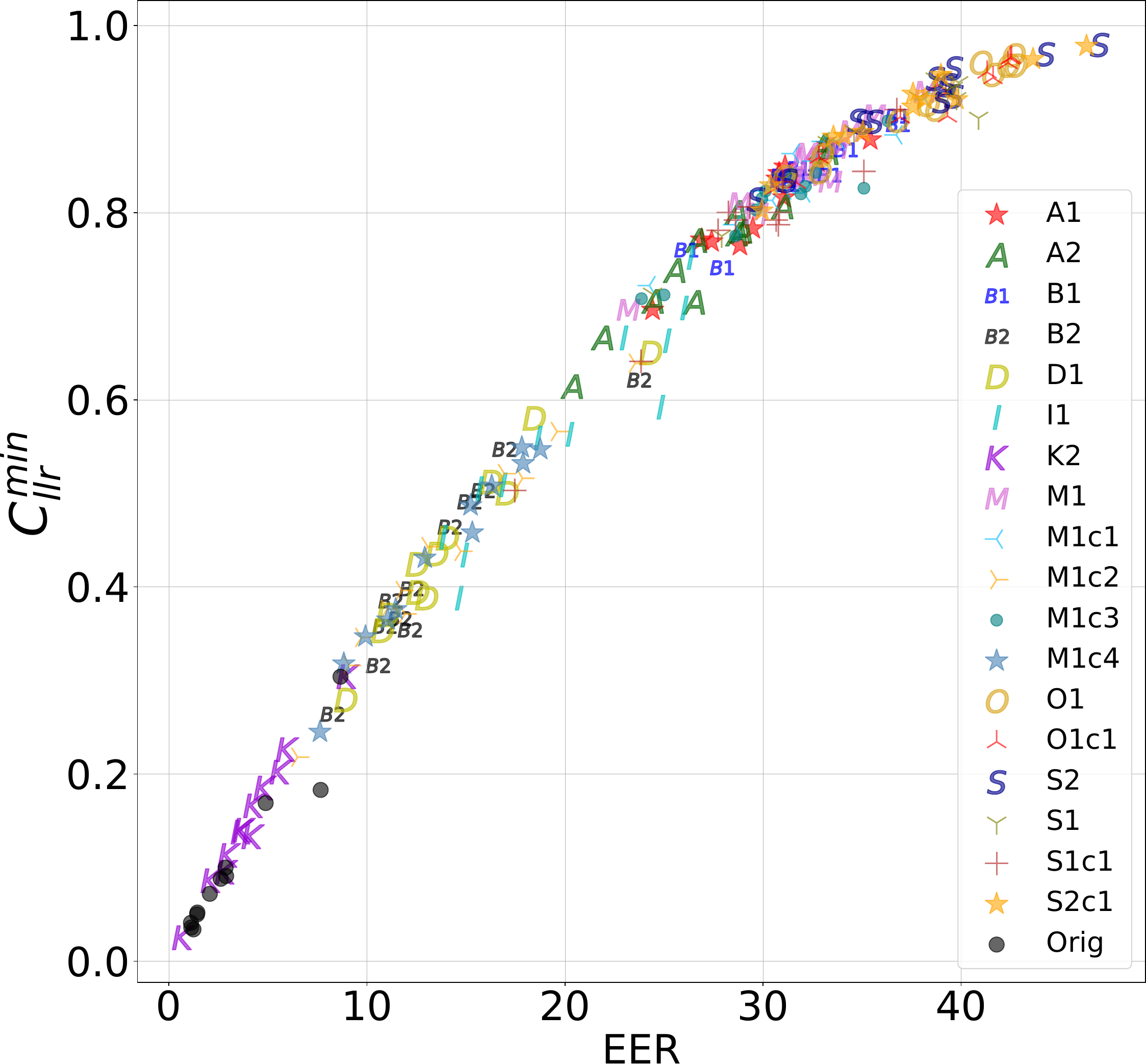}}
\caption{Lazy-informed attacker (aa)} 
\label{fig:fig:eer-vs-cllr-min-aa}
\end{subfigure}
\end{center}
\caption{EER vs.\ \bm{$C_\text{llr}^\text{min}$} results
for the ignorant and lazy-informed attack models.
Each point corresponds to the result of one anonymization system on
 one dataset among the 12 development and evaluation datasets.  Higher EER and $C^\text{min}_\text{llr}$ correspond to better privacy.}
\label{fig:eer-vs-cllr-min}
\end{figure}

The results for other privacy metrics are consistent with those for the EER. See, for example, Figure~\ref{fig:eer-vs-cllr-min} which illustrates EER vs.\  $C^\text{min}_\text{llr}$ results
for  ignorant and lazy-informed attack models for different datasets and systems.
Due to space constrains, we therefore focus on the EER in the following. Results for other metrics are reported by 
\citet[Section~3]{Tomashenko2021CSlsupplementay}.

\begin{figure}[H]
\centering\includegraphics[width=\textwidth]{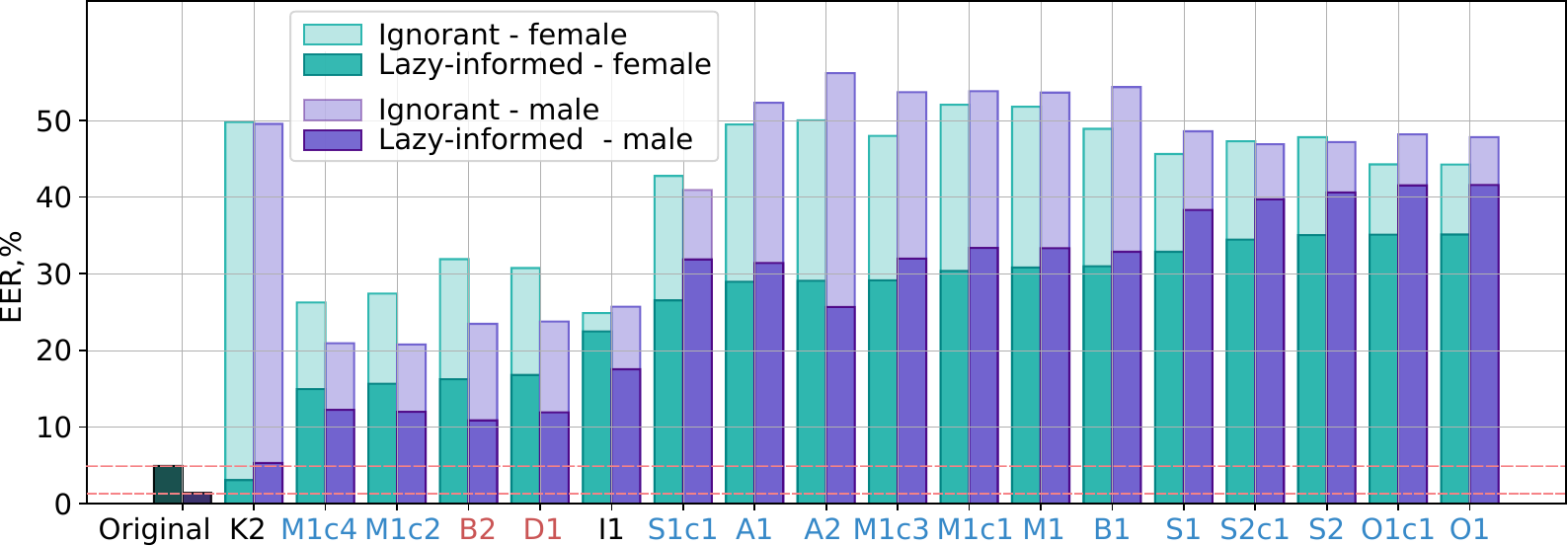}
\caption{Average EER over all test datasets for all anonymization systems and for original data, depending on the attack model and the original speaker's gender.}.
\label{fig:eer-mean-gender}
\end{figure}

Anonymization performance differs at the gender level. Gender-dependent results averaged over all datasets in both evaluation conditions are given in Figure~\ref{fig:eer-mean-gender}.
On the original data, the EER is lower for male speakers than for female speakers.
With only few exceptions (e.g., \textbf{A2}), the opposite is observed after x-vector based anonymization.
Systems \textbf{M1c2} and \textbf{M1c4}, for which the SS AM is fed with original x-vectors, are two of the exceptions, indicating that
gender-dependent differences are the result of x-vector anonymization rather than any extraneous influence, e.g., acoustic mismatch between original and synthesized data.
In contrast, signal processing based approaches show the same gender-dependent trend as the original data.

\subsubsection{Utility: speech recognition error}\label{sec:obj_results_util}

\begin{figure}[H] 
\begin{center}
\centering\includegraphics[width=\textwidth]{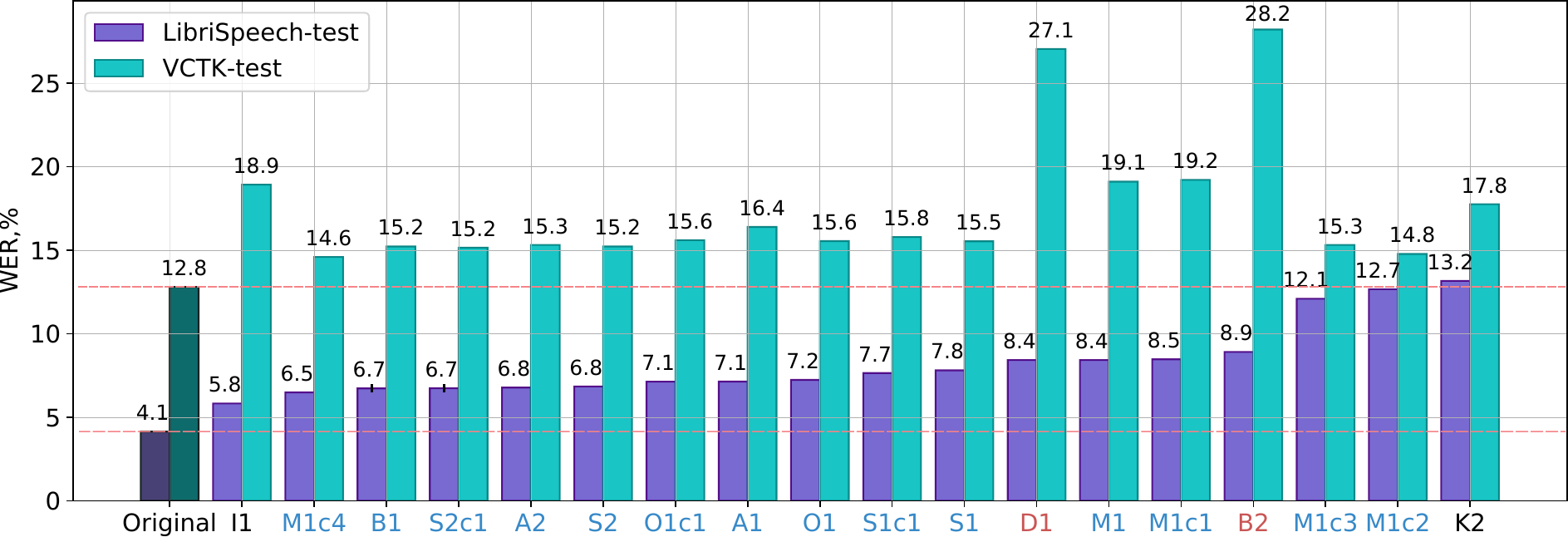}
\caption{WER  on \textit{LibriSpeech-test} and \textit{VCTK-test}  for  all anonymization systems and for original data. Lower WER corresponds to better utility. }\label{fig:wer}
\end{center}
\end{figure}

Figure~\ref{fig:wer} shows the ASR performance in terms of the WER.  Since we observed substantial disparities in the results, they are illustrated separately for the \textit{LibriSpeech-test} and \textit{VCTK-test} datasets.
The WER on original data (no anonymization) is 4.14\% and 12.81\%, respectively.
This difference in the WER is explained as follows:
with the ASR system being trained on the \emph{LibriSpeech-train-clean-360} dataset, performance is better on the matched \emph{LibriSpeech-test} dataset than on the mismatched \emph{VCTK-test} set.

All anonymization systems degrade the WER. In other words, any improvement in privacy comes at the expense of lower utility.
The relative WER increase is more substantial on the \textit{LibriSpeech-test} dataset ($40$--$217$\%) than on the \textit{VCTK-test} dataset ($14$--$120$\%).

 After anonymization, the best WER of 5.83\% on the \textit{LibriSpeech} dataset is obtained by the signal processing based system \textbf{I1}.
Compared to other systems, however, it performs poorly on the \textit{VCTK-test} dataset.  Other signal processing based systems based upon baseline \textbf{B2} fair even worse on this dataset.
On average, on both test sets, x-vector based anonymization techniques related to the primary baseline (\textbf{B1}, \textbf{S2c1}, \textbf{A2}, \textbf{S2}) obtain better results than other systems (and very close to each other). 

Of note is the high WER of system \textbf{M1c2}, which retains the original x-vectors, on the \emph{LibriSpeech-test} dataset. Systems \textbf{M1c3} and \textbf{M1c4}, which partially retain the original x-vectors, yield a higher WER than original data on that dataset too. 
This suggests that resynthesis by itself significantly degrades ASR performance.
The results for systems \textbf{M1} and \textbf{M1c1} (vs.\ \textbf{B1}) indicate that using an end-to-end ASR AM for BN feature extraction degrades ASV performance on both datasets.
For signal processing based techniques (\textbf{I1}, \textbf{D1}, \textbf{B2}) the relative WER degradation is similar across the datasets, while for x-vector based techniques it is much larger on in-domain data with respect to the data used to train the ASR model (\textit{LibriSpeech}) than on out-of-domain data.

\subsubsection{Using anonymized speech data to assess privacy}\label{sec:obj_results_post}

The results reported in Section \ref{sec:obj_results_off} were obtained using an ASV system trained on original data.
We now report evaluation results using ASV systems trained on anonymized data, according to the semi-informed attacker scenario in Section~\ref{sec:obj_eval_metr}.
Four teams submitted anonymized \textit{LibriSpeech-train-clean-360} training data for their primary systems \textbf{O1}, \textbf{M1}, \textbf{S2}, and \textbf{K2}, and we trained four new corresponding $ASV_\text{eval}^\text{anon}$ models on this data. In addition, we trained two $ASV_\text{eval}^\text{anon}$ models on the training data anonymized by the baseline systems \textbf{B1} and \textbf{B2}.
Models were trained in the same way as before, and have the same topology as the $ASV_\text{eval}$ model trained on original data.

\begin{figure}[H]
\begin{center}
\center{\includegraphics[width=80mm]{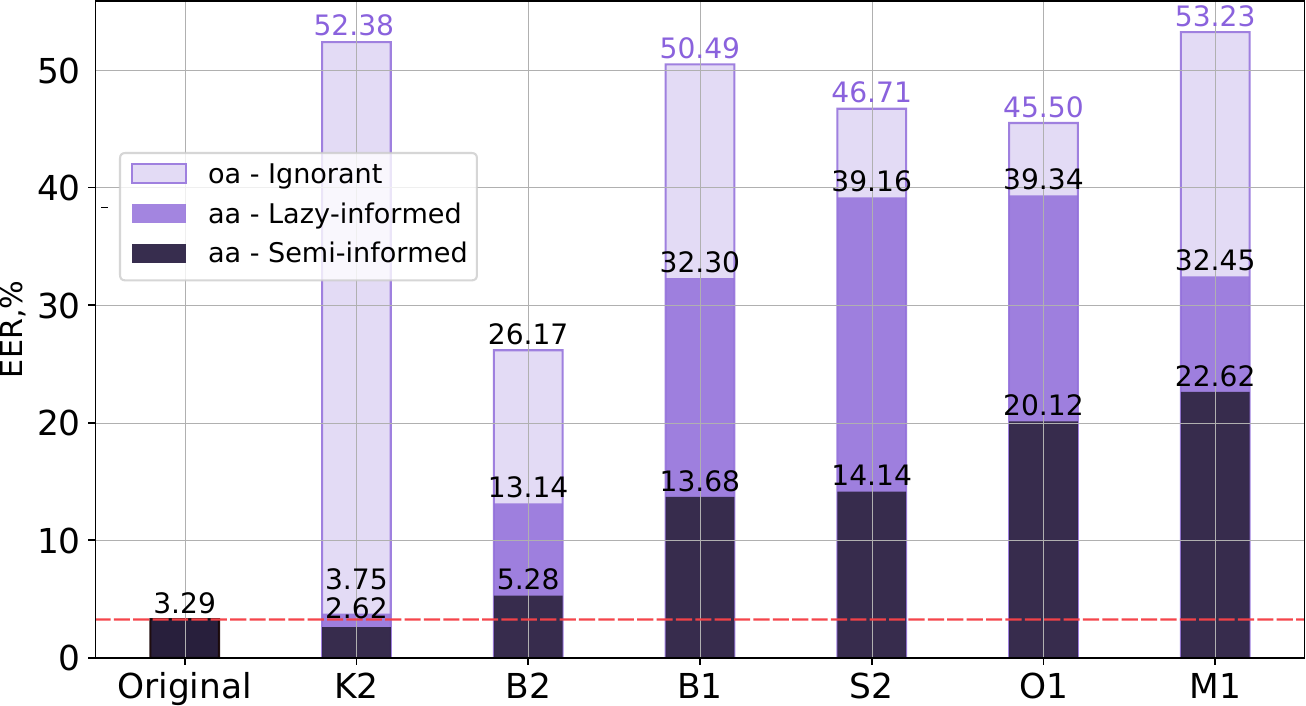}}
\end{center}
\caption{Average EER over all test datasets for a subset of anonymization systems 
and for original data, against the three attack models.
}
\label{fig:anon-eer-all}
\end{figure}

Figure~\ref{fig:anon-eer-all} compares the average EERs obtained for the semi-informed (dark, lower bars), lazy-informed, and ignorant attack models.
For all anonymization systems, training the ASV evaluation model on anonymized data significantly decreases the EER: EERs are substantially lower against semi-informed than ignorant or lazy-informed attackers. 
Thus, assessing the performance of anonymization systems using an ASV system trained on original data leads to a false impression of protection; if the ASV system is retrained on similarly anonymized data, the level of protection becomes closer to (but still better than) that for original, unprotected data.

\subsubsection{Using anonymized speech data to assess utility }\label{sec:obj_results_post_asr}

\begin{figure}[H] 
\centering\includegraphics[width=80mm]{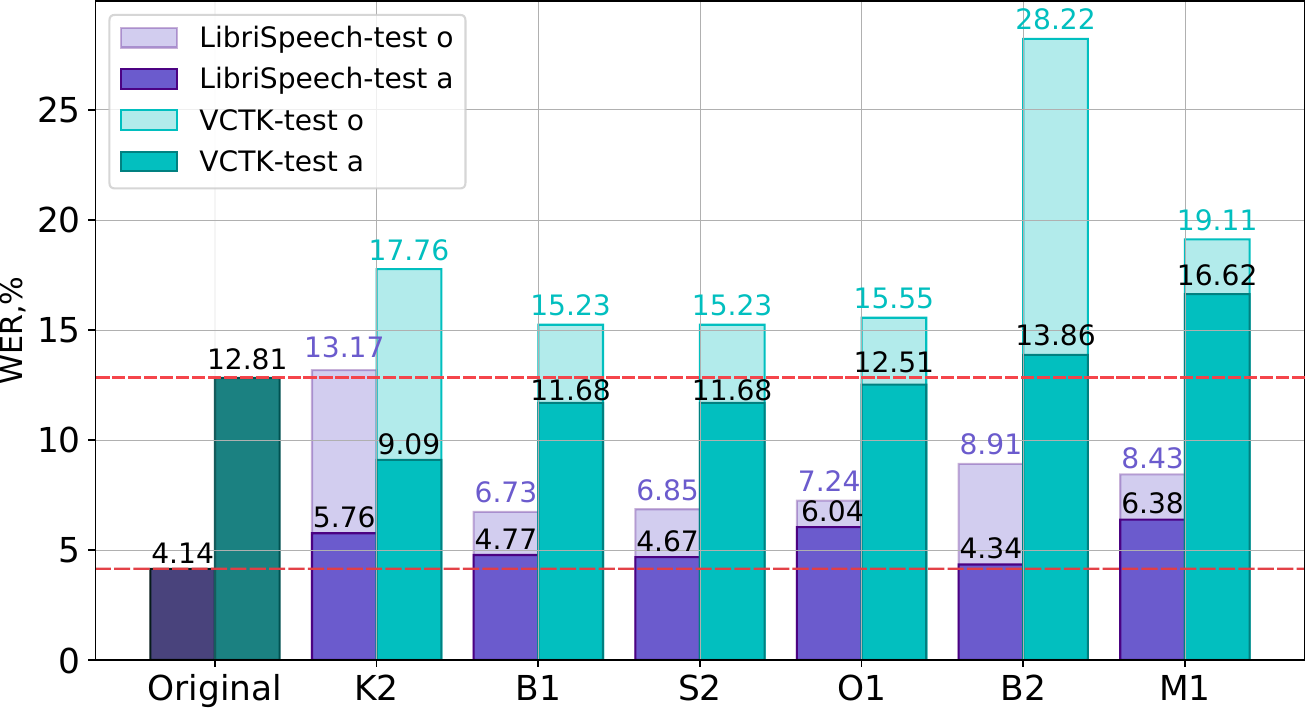}
\caption{WER on \textit{LibriSpeech-test} and \textit{VCTK-test} for a subset of anonymization systems and for original data, evaluated using $ASR_\text{eval}$ (\textbf{o}) or $ASR_\text{eval}^\text{anon}$ (\textbf{a}).
} 
\label{fig:wer-posteval}
\end{figure}

Similarly, Figure~\ref{fig:wer-posteval} compares the WERs obtained by ASR systems trained on anonymized speech data ($ASR_\text{eval}^\text{anon}$) with those obtained by the ASR system trained on original data ($ASR_\text{eval}$). 
The WERs for $ASR_\text{eval}^\text{anon}$ (dark, lower bars, \textbf{a}) are consistently lower than for $ASR_\text{eval}$ (light, upper bars, \textbf{o}).
In some cases, the WER decreases to a level close to that of $ASR_\text{eval}$ on original data.
This implies that degradations to utility can be offset simply by retraining using similarly anonymized data. 
This substantially improves the trade-off between privacy and utility; there is potential to protect privacy with only modest impacts upon utility.

\subsubsection{De-identification and gain of voice distinctiveness }\label{sec:voice_dist}
\begin{figure}[H]
\begin{center}
\begin{subfigure}{0.158\textwidth}
\center{\includegraphics[width=1\textwidth]{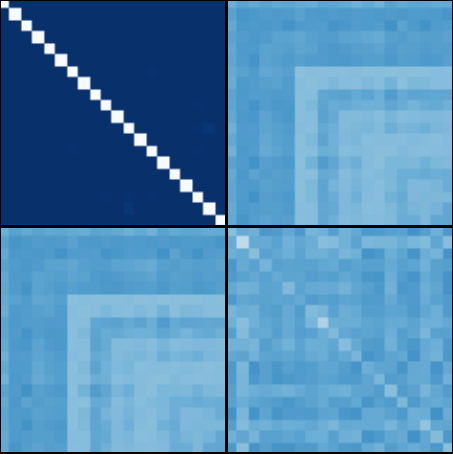}}
\caption{\textcolor{frenchblue}{\textbf{B1}}}
\label{fig:mat-libri-f-b1}
\end{subfigure}
\begin{subfigure}{0.158\textwidth}
\center{\includegraphics[width=1\textwidth]{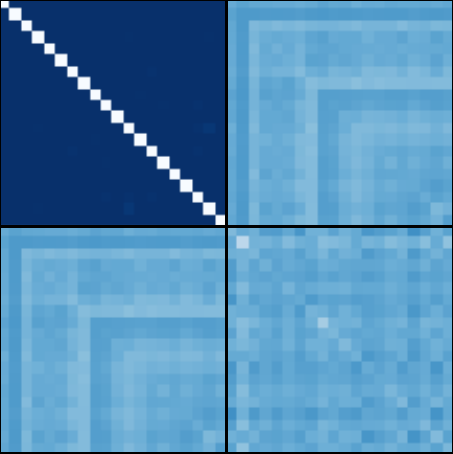}}
\caption{\textcolor{frenchblue}{\textbf{S2}}} 
\label{fig:mat-libri-f-s2}
\end{subfigure}
\begin{subfigure}{0.158\textwidth}
\center{\includegraphics[width=1\textwidth]{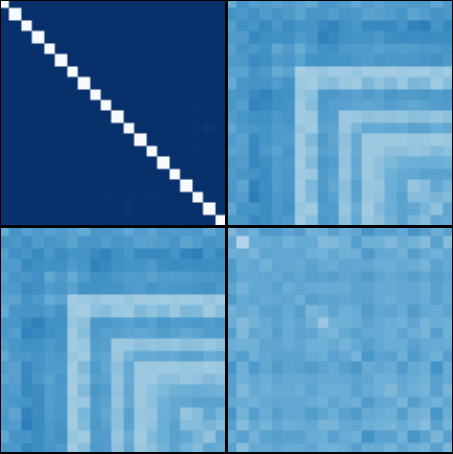}}
\caption{\textcolor{frenchblue}{\textbf{O1}}} 
\label{fig:mat-libri-f-o1}
\end{subfigure}
\begin{subfigure}{0.158\textwidth}
\center{\includegraphics[width=1\textwidth]{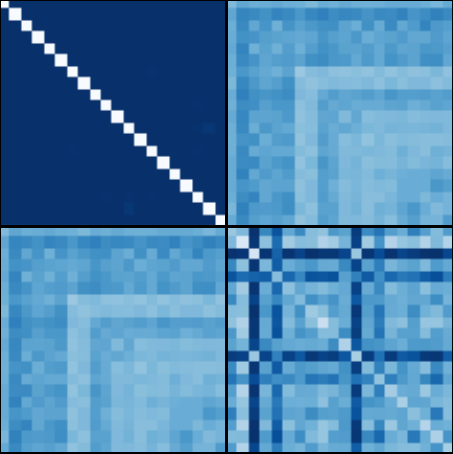}}
\caption{\textcolor{frenchblue}{\textbf{S1}}} 
\label{fig:mat-libri-f-s1}
\end{subfigure}
\begin{subfigure}{0.158\textwidth}
\center{\includegraphics[width=1\textwidth]{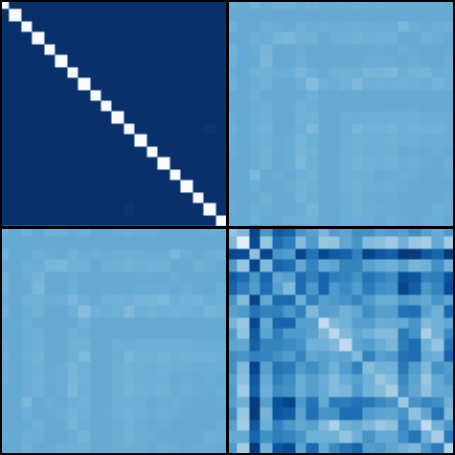}}
\caption{\textcolor{frenchblue}{\textbf{A1}}} 
\label{fig:mat-libri-f-a1}
\end{subfigure}
\begin{subfigure}{0.158\textwidth}
\center{\includegraphics[width=1\textwidth]{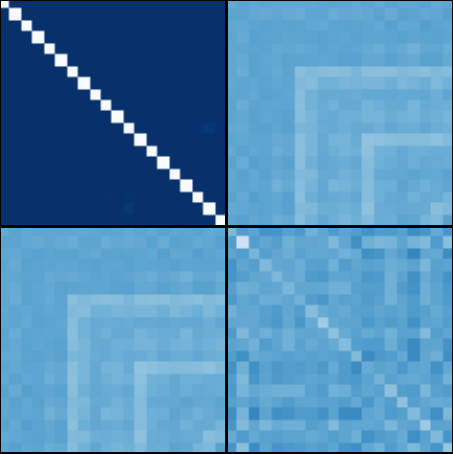}}
\caption{\textcolor{frenchblue}{\textbf{M1}}} 
\label{fig:mat-libri-f-m1}
\end{subfigure}
\begin{subfigure}{0.158\textwidth}
\center{\includegraphics[width=1\textwidth]{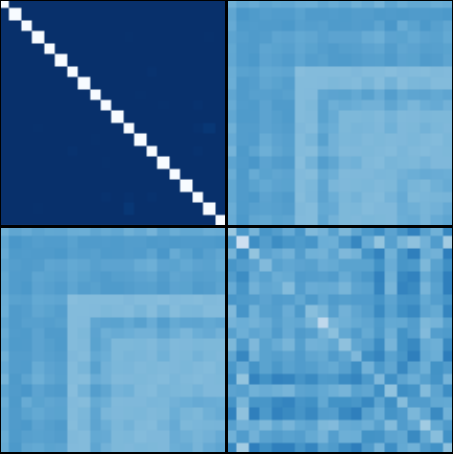}}
\caption{\textcolor{frenchblue}{\textbf{A2}}}
\label{fig:mat-libri-f-a2}
\end{subfigure}
\begin{subfigure}{0.158\textwidth}
\center{\includegraphics[width=1\textwidth]{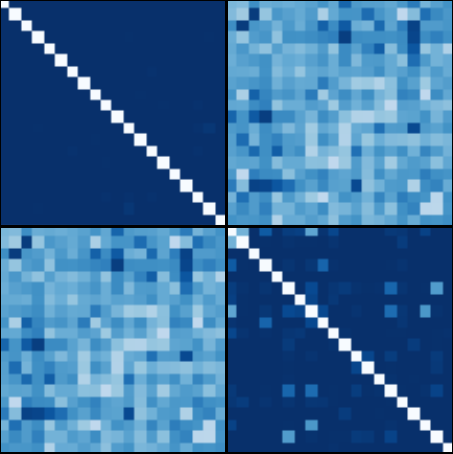}}
\caption{K2} 
\label{fig:mat-libri-f-k2}
\end{subfigure}
\begin{subfigure}{0.158\textwidth}
\center{\includegraphics[width=1\textwidth]{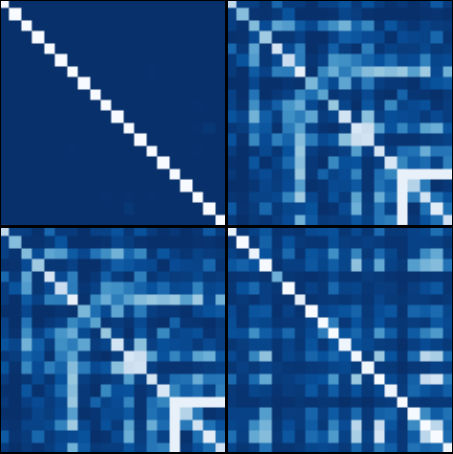}}
\caption{\textcolor{chestnut}{\textbf{B2}}}
\label{fig:mat-libri-f-b2}
\end{subfigure}
\begin{subfigure}{0.158\textwidth}
\center{\includegraphics[width=1\textwidth]{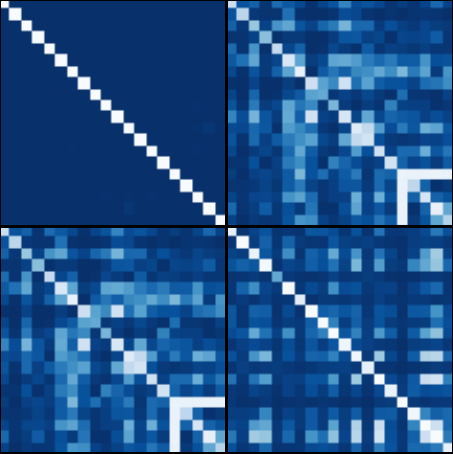}}
\caption{\textcolor{chestnut}{\textbf{D1}}} 
\label{fig:mat-libri-f-d1}
\end{subfigure}
\begin{subfigure}{0.191\textwidth}
\center{\includegraphics[width=1\textwidth]{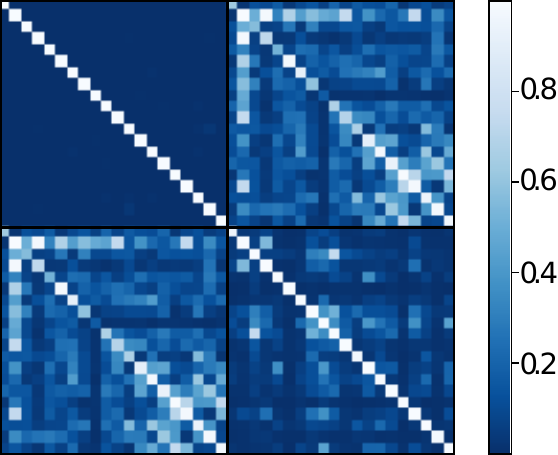}}
\caption{I1} 
\label{fig:mat-libri-f-i1}
\end{subfigure}
\end{center}
\caption{Voice similarity matrices for all primary systems on the female speakers of the \textit{LibriSpeech-test} dataset. The global matrix $M$ for each system is composed of the three submatrices $M_\text{oo}$, $M_\text{oa}$ and $M_\text{aa}$ defined in Section~\ref{sec:obj_eval_metr} as {$M=\tiny\begin{pmatrix}
M_\text{oo} & M_\text{oa} \\
M_\text{oa} & M_\text{aa}
\end{pmatrix}$.} }
\label{fig:mat-libri-f-x}
\end{figure}

Figure~\ref{fig:mat-libri-f-x} illustrates the voice similarity matrices obtained for all primary systems.
The distinct diagonal in $M_\text{oo}$ (top left submatrix of each matrix $M$) points out the speaker discrimination ability in the original data. The two other submatrices, $M_\text{oa}$ (top right) and $M_\text{aa}$ (bottom right), show substantial differences across the systems. 
In $M_\text{oa}$ the diagonal disappears if \added{the pseudo-speakers differ from the original speakers}, while in $M_\text{aa}$ the diagonal emerges if the pseudo-speakers can be distinguished from each other \cite{noe2020speech}. 
The matrices for signal processing based systems and for system \textbf{K2} exhibit a distinct diagonal in $M_\text{aa}$, indicating that voices remain distinguishable after anonymization. For x-vector based systems, this diagonal is much weaker.  

\begin{figure}[H]
\begin{center}
\begin{subfigure}{0.47\textwidth}
\center{\includegraphics[width=1\textwidth]{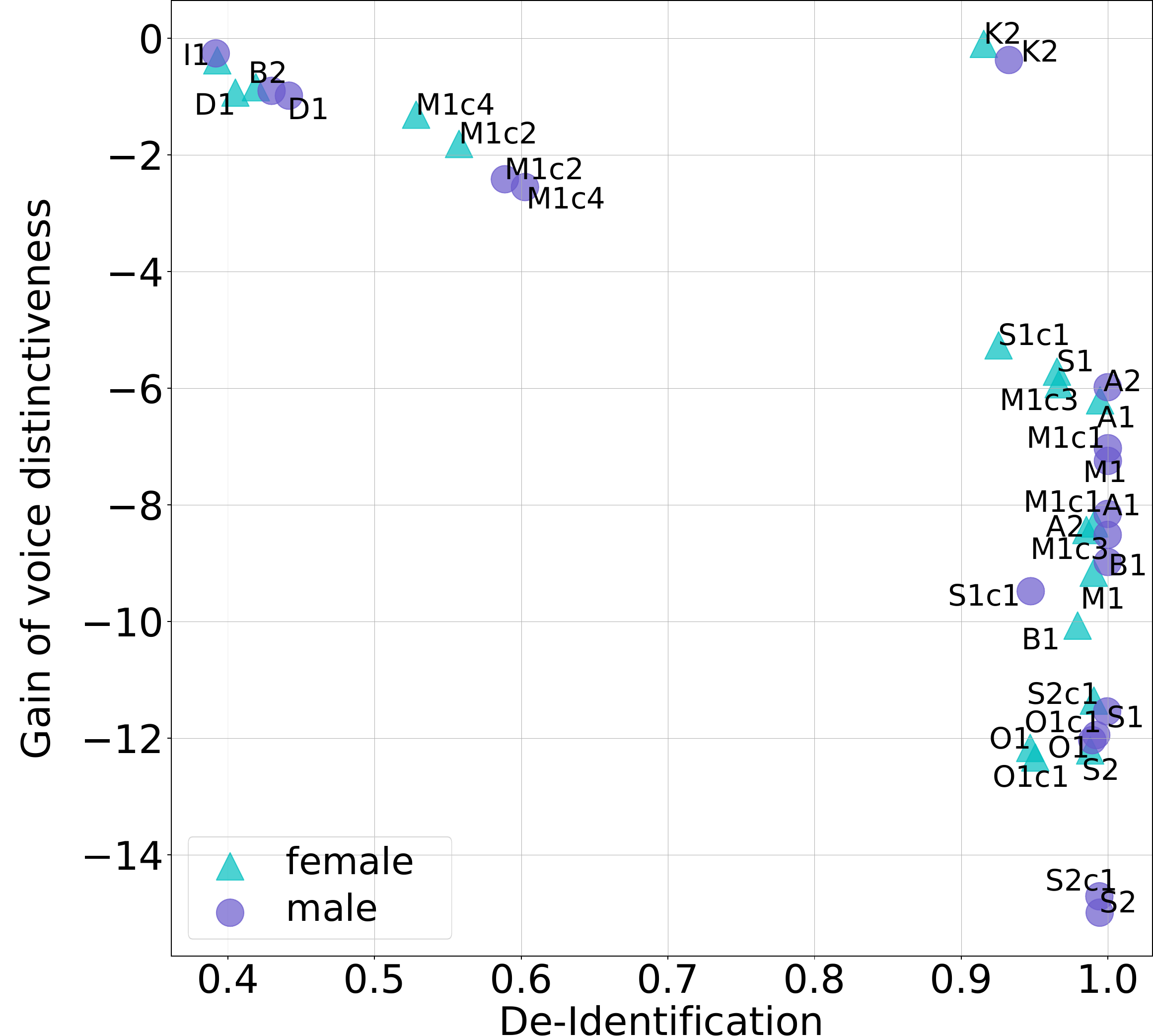}}
\caption{LibriSpeech-test}
\end{subfigure}
~~~
\begin{subfigure}{0.47\textwidth}
\center{\includegraphics[width=1\textwidth]{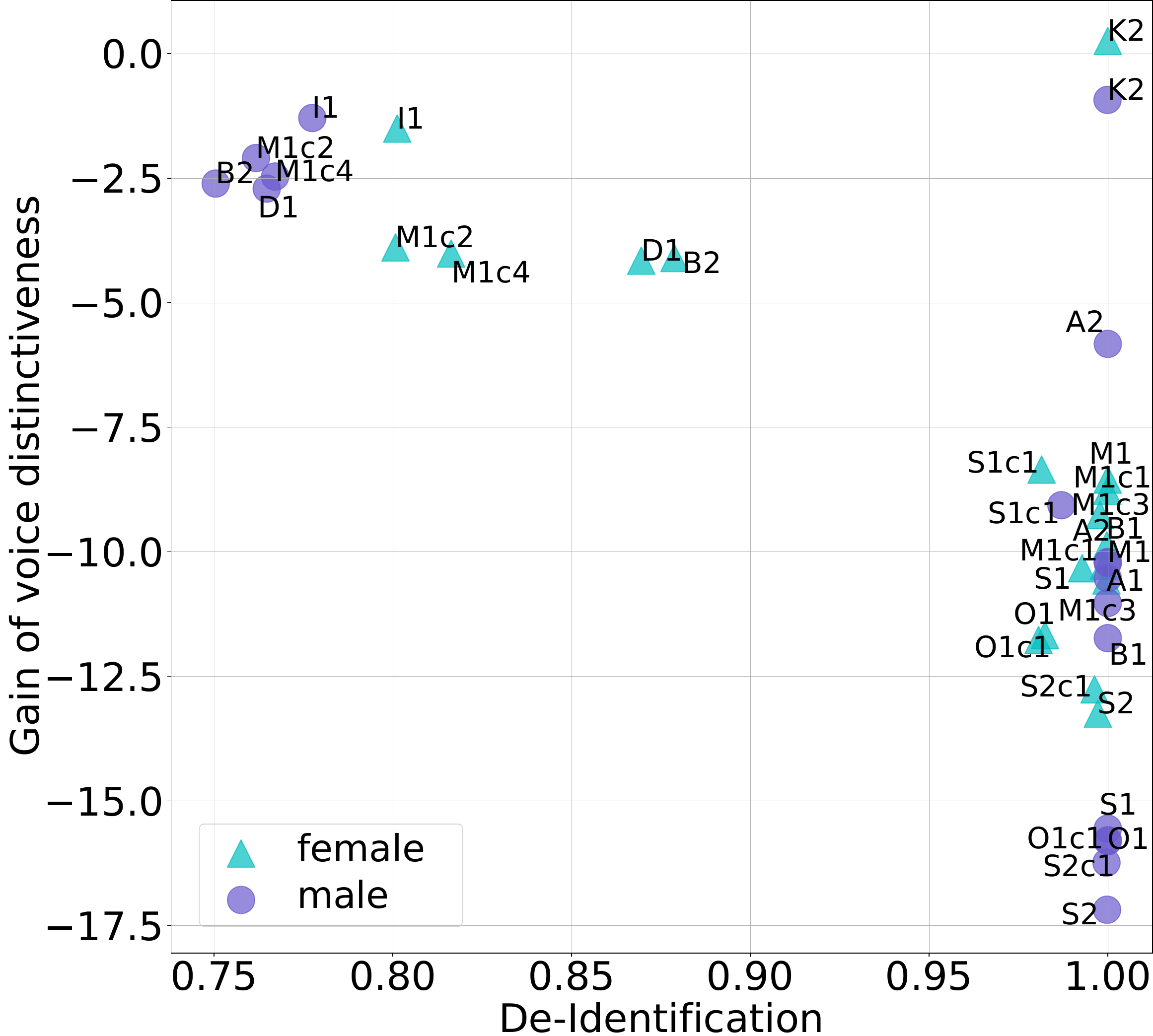}}
\caption{VCTK-test (different)} 
\end{subfigure}
\end{center}
\caption{De-identification (DeID) vs.\ gain of voice distinctiveness ($G_{\text{VD}}$) for all anonymization systems.
Higher DeID corresponds to better privacy, higher $G_{\text{VD}}$ to better distinctiveness of anonymized voices. }
\label{fig:deid-vd}
\end{figure}

The scatter plots in Figure~\ref{fig:deid-vd} show the \emph{gain of voice distinctiveness} ($G_{\text{VD}}$) against \emph{de-identification} performance (DeID) for the \emph{LibriSpeech-test} (left) and \emph{VCTK-test} (right) datasets.\footnote{For more details, see 
\citet[Sections~3.4 and 3.5]{Tomashenko2021CSlsupplementay}}
The results show that systems based upon baseline \textbf{B1} provide close to perfect de-identification, while signal processing based solutions tend to better preserve voice distinctiveness. 
For the latter, de-identification performance varies across the datasets.
Only system \textbf{K2} achieves high de-identification with only modest degradation to voice distinctiveness.
The results for systems \textbf{M1c4} and \textbf{M1c2} which use original x-vectors show that copy-synthesis alone also degrades voice distinctiveness. Interestingly, de-identification performance for both systems is comparable to that for signal-processing based methods.  These observations are consistent with EER and $C^\text{min}_\text{llr}$ results.

The results in Figure~\ref{fig:deid-vd} also show that different systems lead to differences in voice distinctiveness for different genders. In particular, systems \textbf{S2} and \textbf{S2c1} better preserve distinctiveness for female speakers, while system \textbf{A2} better preserves distinctiveness for male speakers.

\subsubsection{Relation between privacy and utility metrics}

\begin{figure}[H]
\begin{center}
\begin{subfigure}{0.47\textwidth}
\center{\includegraphics[width=1\textwidth]{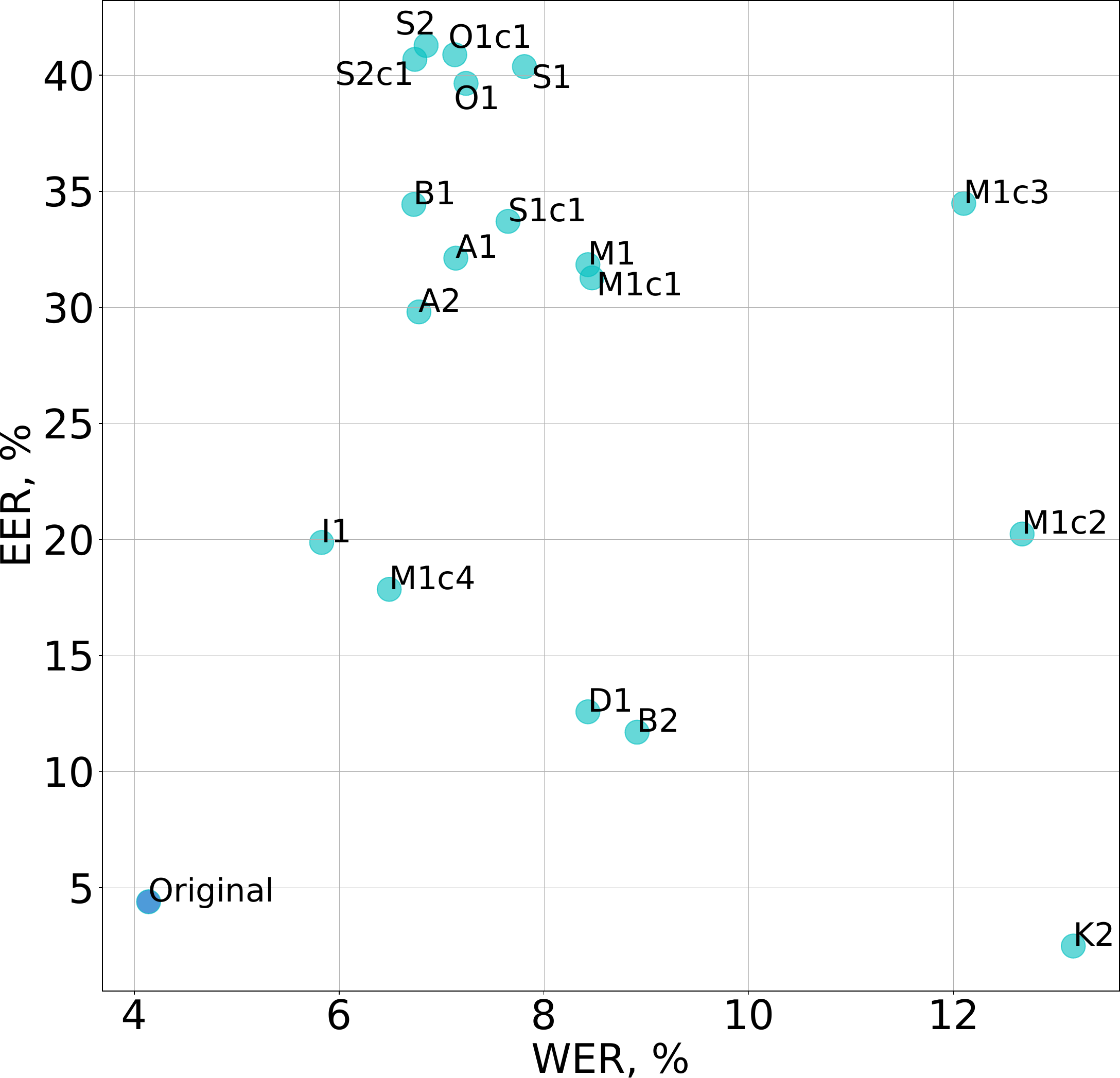}}
\caption{LibriSpeech-test}
\label{fig::wer-eer-libri}
\end{subfigure}
~~~
\begin{subfigure}{0.47\textwidth}
\center{\includegraphics[width=1\textwidth]{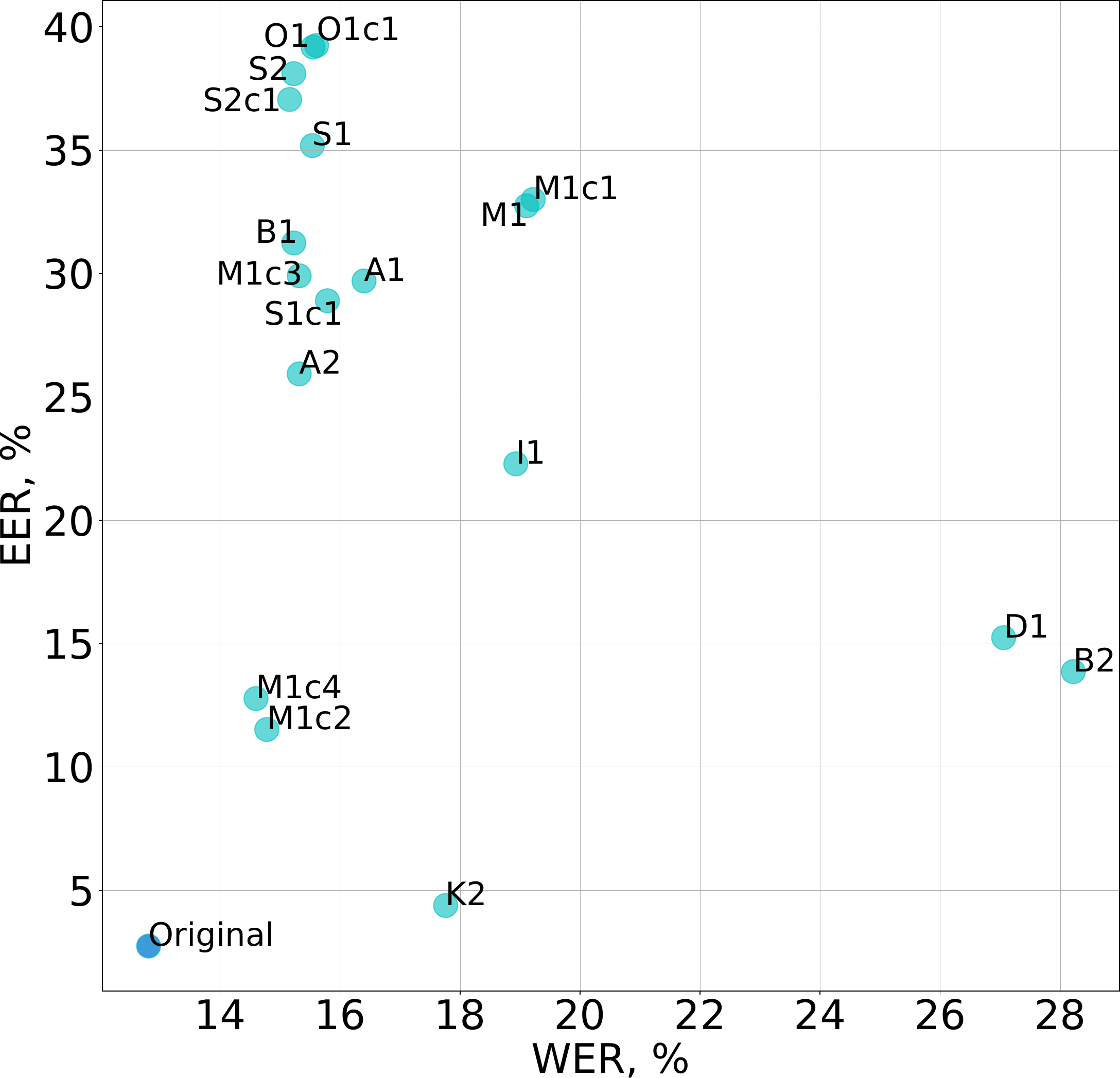}}
\caption{VCTK-test} 
\label{fig::wer-eer-vctk}
\end{subfigure}
\end{center}
\caption{WER vs.\ EER on \textit{LibriSpeech-test} and \textit{VCTK-test} for all anonymization systems and for original data, evaluated using $ASR_\text{eval}$ and the lazy-informed attack model.}
\label{fig:wer-eer}
\end{figure}

As we observed above, all anonymization systems reduce the utility of  speech data. 
Therefore, it is  important to consider the trade-off between privacy and utility.
Figure~\ref{fig:wer-eer} demonstrates the relation between objective privacy and utility in the form of scatter plots with WER and EER values for all anonymization systems\added{, evaluated using $ASR_\text{eval}$ and $ASV_\text{eval}$ systems trained on original data}.
The best anonymization system should have maximum EER and minimum WER, i.e., be close to the top-left corner.
We can see that there is no system which provides the best results for both metrics.
On the \textit{LibriSpeech-test} dataset, the best anonymization is achieved using x-vector based systems, while the lowest WER corresponds to system \textbf{I1} which is a signal processing based system. However on \textit{VCTK-test}, the results for this system are different and better results for both metrics are obtained using x-vector based systems.

\subsection{Subjective evaluation results}\label{sec:subj_results}
This section presents subjective evaluation results for speech naturalness, intelligibility, and speaker verifiability (Sections~\ref{sec:res_subj_violin} and \ref{sec:res_subj_det}), and speaker linkability (Section~\ref{sec:res_subj_link}).

\subsubsection{Distribution of naturalness, intelligibility, and verifiability scores}\label{sec:res_subj_violin}
\begin{figure}[t!]
\centering\includegraphics[width=\textwidth]{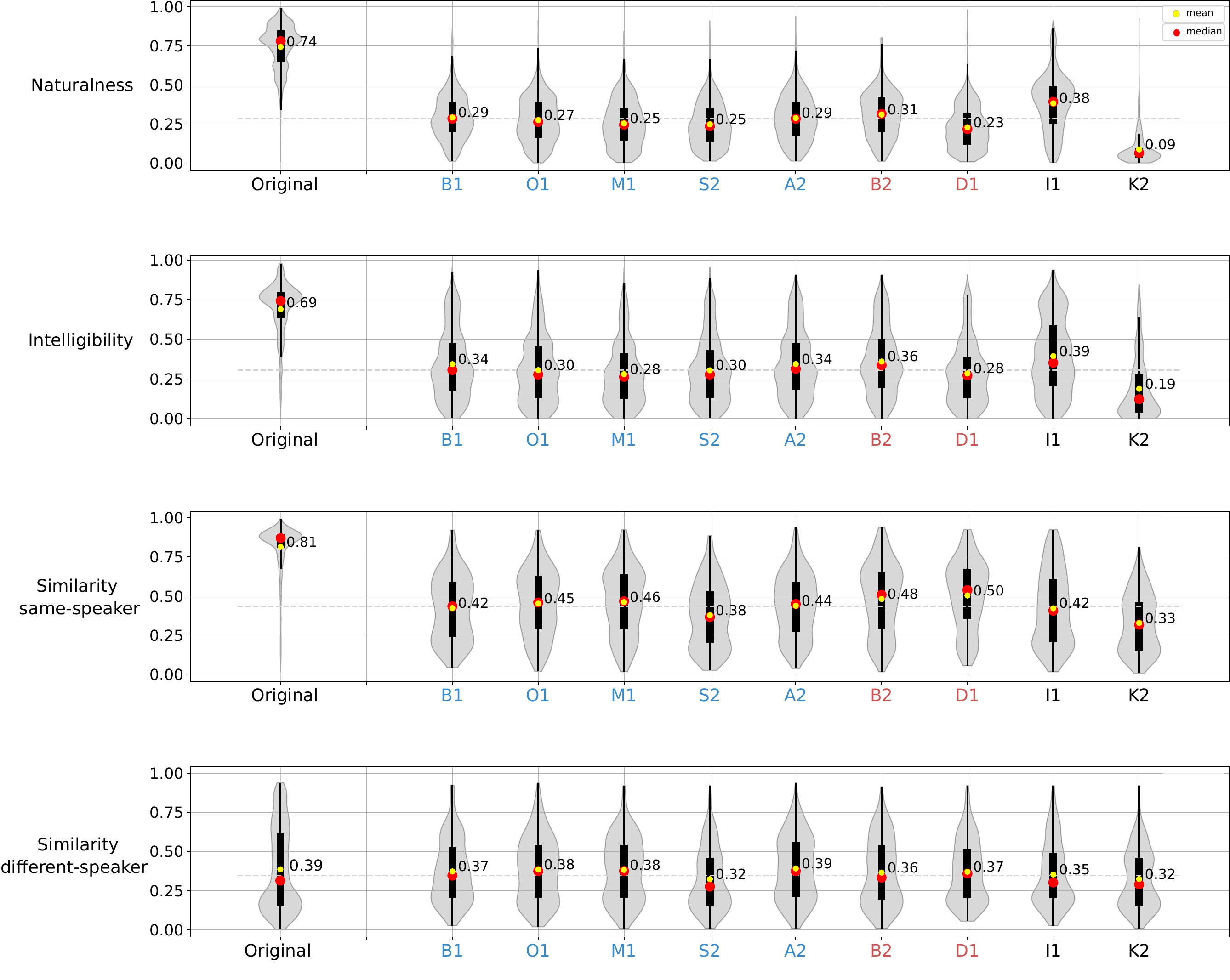}
\caption{Violin plots of normalized {\textbf{subjective speech naturalness, intelligibility}}, and {\textbf{speaker similarity}} scores \replaced{pooled over \textit{LibriSpeech-test} and \textit{VCTK-test}.}{For naturalness and intelligibility,
scores from target and non-target anonymized data are pooled; for similarity, scores for anonymized target and non-target speakers data are separately
plotted in 3rd and 4th sub-figures, respectively.} The dotted line indicates the median for \textbf{B1}. Numbers indicate mean values. Higher naturalness and intelligibility scores correspond to better utility, and lower similarity scores to better privacy. 
}
\label{fig:bar-norm}
\end{figure}

The distributions of normalized naturalness, intelligibility, and speaker similarity scores obtained from the unified subjective test are displayed in Figure~\ref{fig:bar-norm} as violin plots \cite{hintze1998violin}.\footnote{Statistical significance test results are reported by 
\citet[Tables~16 and~17]{Tomashenko2021CSlsupplementay}.} 
The similarity scores for same-speaker and different-speaker pairs are plotted separately, since they are expected to be different.

The results for naturalness and intelligibility are as expected. Anonymized samples from all systems are inferior to the \deleted{target and non-target} original data, and the differences are statistically significant at $p\ll 0.01$. This performance gap exists in both methods based on the primary baseline  (\textbf{B1}, \textbf{O1}, \textbf{M1}, \textbf{S2}, and \textbf{A2}) and the secondary baseline  (\textbf{B2}, \textbf{D1}). 
While \textbf{I1} outperforms the other anonymization systems in terms of naturalness, it is still far from perfect in terms of both naturalness and intelligibility. More efforts are necessary to address the degradation caused by existing anonymization methods.

Concerning speaker similarity, the anonymized \added{trial} data from \replaced{a given speaker}{target speakers} are perceptually \added{much} less similar to the \added{original} enrollment data of \replaced{that}{the target} speaker than the \replaced{original trial data of that}{unanonymized trial of that target} speaker. This indicates that all systems achieve a \replaced{good}{certain} degree of anonymization according to human perception.

\subsubsection{Naturalness, intelligibility, and verifiability DET curves}\label{sec:res_subj_det}
To further investigate the difference across systems, we plot
\added[id=rev, comment=1.15]{detection error trade-off (DET) curves \cite{martin1997det}. These curves assume a detection task, where the decision for a given trial is made by comparing the score with a threshold. The false alarm and miss rates are computed as a function of the threshold and plotted against each other.
For naturalness and intelligibility the task is to detect original data, while for speaker similarity the task is to detect whether the trial utterance is from the same speaker as the enrollment utterance. The closer the DET curves are to the top-right corner of each plot, the higher the naturalness, intelligibility, and privacy preservation. Once again, the DET curves for same-speaker and different-speaker pairs are plotted separately, since they are expected to be different.} 
\deleted{Since there are four types of scores, i.e., \{target original, non-target original, target anonymized, non-target anonymized\}, we computed the DET curves in the following ways:}

\deleted{For naturalness and intelligibility, an ideal anonymization system should have a DET curve close to that of original data, indicating similar  naturalness and intelligibility scores to the original data and therefore minimum degradation on naturalness and intelligibility. For similarity curve 1, an ideal anonymization system should have a DET curve close to the diagonal line from bottom-right to top-left, indicating that the anonymized data of a target speaker sounds similar to the non-target data.}

\begin{figure}[t!]
\centering\includegraphics[width=\textwidth]{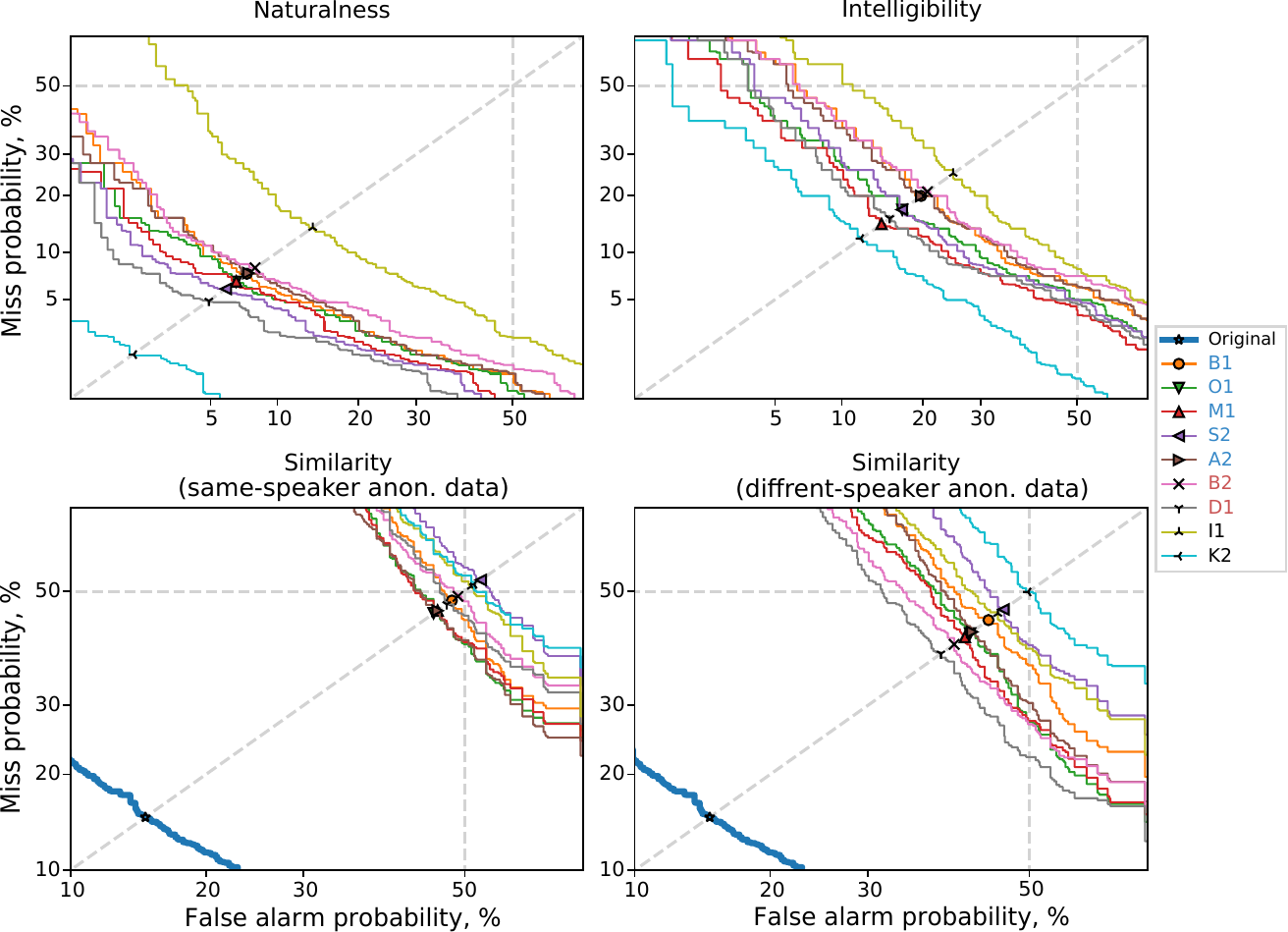}
\caption{DET curves based on subjective evaluation scores pooled over  \textit{LibriSpeech-test} and \textit{VCTK-test}. 
}
\label{fig:det-pool-norm}
\end{figure}

The four types of DET curves are plotted in Figure~\ref{fig:det-pool-norm}.\footnote{For separate results over \textit{LibriSpeech-test} and \textit{VCTK-test}, see 
\citet[Figure~26]{Tomashenko2021CSlsupplementay}.}
Concerning naturalness and intelligibility, \deleted{the DET curves of the original data are straight lines across the top-right corner, indicating that the scores of non-target original data are similar to those of the target original data. This is expected because original data should have similar naturalness and intelligibility no matter whether they are from target or non-target speakers. In contrast,} the DET curves for anonymized data are far from the \replaced{top-right corner}{curve of original data}, suggesting that anonymized data are inferior to original data in terms of naturalness and intelligibility.
The naturalness DET curves of \textbf{I1} and \textbf{K2} seem to deviate from the other anonymization systems. While other systems are based on either \textbf{B1} or \textbf{B2}, \textbf{I1} uses a different signal processing based approach, and \textbf{K2} uses a different deep learning method. 
As such, \textbf{I1} avoids several errors such as ASR AM errors in \textbf{B1}, which may contribute to its naturalness. However, it is interesting to note how different signal processing algorithms result in different perceptual naturalness and intelligibility. Also note that none of the systems except \textbf{I1} outperforms \textbf{B2}.

Concerning speaker similarity, \replaced{both in the same-speaker and different-speaker cases,}{let us focus on the case where target speaker data is anonymized (left-bottom figure in~\ref{fig:det-pool-norm}). We observe that} the DET curves of original data are close to the bottom-left corner while those of anonymized data are close to the top-right corner. In other words, \replaced{anonymization of the trial utterances makes it difficult to decide whether the original enrollment utterance comes from the same speaker or not}{the anonymized data of target speaker produced similar perceptual scores to the non-target speaker data, indicating that anonymized target speaker data sound less similar to the original target speaker}. 
The similarity DET curves of \textbf{K2}, \textbf{S2}, and \textbf{I1} \replaced{in the same-speaker case}{on target speaker data} are closer to the top-right corner than others. However, these three systems behave quite differently in terms of naturalness and intelligibility, with \textbf{I1} and \textbf{K2} achieving the highest and lowest median score, respectively. This implies that an anonymized trial may sound like the voice of a different speaker simply because of the severe distortion caused by anonymization. 

To sum up, all the submitted anonymization systems can conceal the perceived speaker identity to some degree. However, none of them can produce anonymized speech that is as natural and intelligible as original speech. One signal processing based anonymization method (\textbf{I1}) degrades the naturalness and intelligibility less severely, but it still degrades them to some extent.

\subsubsection{Perception of speaker identity and speaker linkability}\label{sec:res_subj_link}

We report speaker linkability results for the two baseline systems in terms of the \textit{F-measure} ($F_1$), \textit{clustering change (CC)}, and \textit{clustering purity} metrics. To measure the effects of anonymization for each evaluator, we calculated the \textit{difference} between the values of the $F_1$ and CC metrics on the control \added{panel} (original data only) and their average values over the two other \added{panels} (half of the data anonymized by \textbf{B1} or \textbf{B2}). 

\begin{figure}[!h]
\centering
\includegraphics[width=12cm]{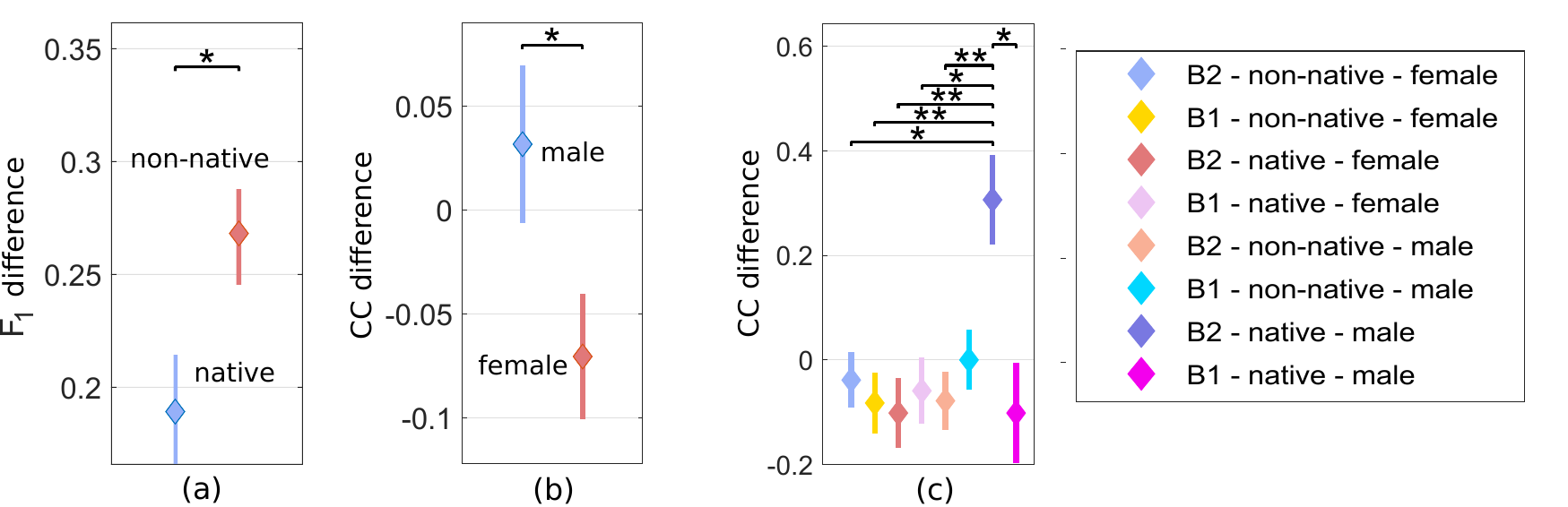}
\caption{Diamonds and vertical lines represent the means and standard errors, respectively. 
(a)~Mean $F_{1}$ difference 
\added{for}
native and non-native English speaking evaluators. \{*\} signifies $p<0.05$.
(b)~Mean CC difference 
depending on the original speaker's gender. 
(c)~Mean CC difference
\added{for}
anonymization system$\times$language$\times$original gender interactions. \{*, **\} signify $p<\{0.05, 0.01\}$.}
\label{anova}
\end{figure}

We observed a main effect on the mean $F_{1}$ difference of the evaluator's native language $F_{1,64}=6.5$, $p<0.05$, $\eta_{p}^2=0.09$, but no effects of the anonymization system nor the original speaker's gender, $p>0.05$. \textbf{B1} evaluators exhibited a greater mean $F_{1}$ difference ($0.24\pm0.02$) than \textbf{B2} evaluators ($0.21\pm 0.02$). Post-hoc t-tests showed that non-native English speaking evaluators were more affected by linking natural and anonymized utterances ($0.26\pm0.02$) than native English speaking evaluators ($0.19\pm 0.022$) (Figure~\ref{anova}a).

For the mean CC difference, we found a main effect of the original speaker's gender $F_{1,64}=4.45$, $p<0.05$, $\eta_{p}^2=0.06$, and interactions for anonymization system $\times$ language $F_{1,64}=4.26$, $p<0.05$, $\eta_{p}^2=0.06$ and anonymization system $\times$ language $\times$ original gender $F_{1,64}= 8.75$, $p<0.01$, $\eta_{p}^2=0.11$. Post-hoc t-tests revealed that evaluators showed a greater mean  CC difference when presented male utterances ($0.07\pm0.03$) in comparison to female ($-0.03\pm0.04$) (Figure~\ref{anova}b). Native English speaking evaluators also exhibited a greater mean CC difference than \textbf{B2} evaluators (Figure~\ref{anova}c). 
These results suggest that the evaluators were able to use the anonymized utterances to aid their performance when grouping female utterances, whereas performance diminished when they listened to anonymized male utterances. Non-native English speaking evaluators achieved a lower accuracy when presented with anonymized stimuli from either system. Overall, the above results suggest  that the perceptual effectiveness of an anonymization system can depend on the users as well as on the attacker (here, the evaluator).

\begin{figure}[H]
\begin{center}
\begin{subfigure}[b]{0.38\textwidth}
\center{\includegraphics[width=1\textwidth]{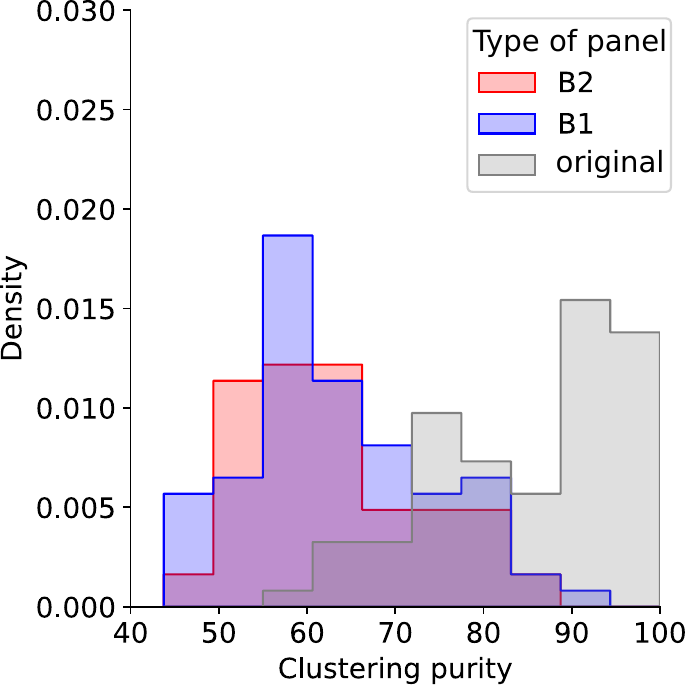}}
\caption{}
\label{fig:cluster-purity}
\end{subfigure}
~~~
\begin{subfigure}[b]{0.38\textwidth}
\center{\includegraphics[width=1.2\textwidth]{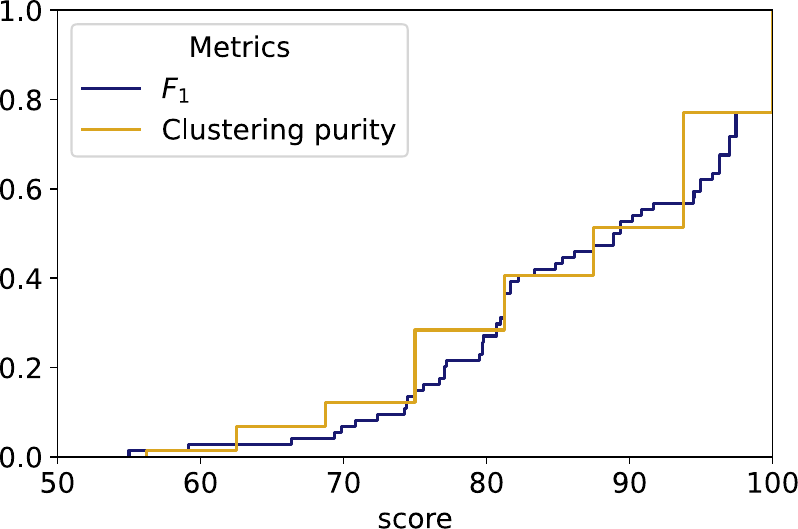}}
\caption{} 
\label{fig:cum-orig}
\end{subfigure}
~~~~
\end{center}
\caption{ (a) density distributions for clustering purity; (b) cumulative density for  clustering purity and $F_1$ on the original (control) trials.
}
\label{fig:clust}
\end{figure}

The distribution of clustering purity for the three \added{panels}
is displayed in Figure~\ref{fig:cluster-purity}. The Mann-Whitney test shows an effect of the \added{panel} (control vs.\ other) on the purity: $\chi^2=\numprint{82688}$ ($p<0.001$) for female speakers and  $\chi^2=\numprint{41344}$ ($p<0.001$) for male speakers, which indicates that the distributions for the original and the anonymized \added{panels} are different. As expected, the evaluators achieve a higher average purity ($86.40\%$) on the original \added{panel} than on the two other \added{panels} ($61.68\%$ and $62.58\%$).
These results indicate that linking an anonymized voice to its original counterpart is not as easy as clustering original voices.
 The distribution of the clustering purity is similar to that of $F_1$ for all \added{panel} types (see Figure~\ref{fig:cum-orig}).
No significant difference between the two baselines is noticed for both metrics.

\normalsize

\subsection{Comparison of objective and subjective evaluation results}\label{sec:compare_results}

\begin{figure}[h!]
\center{\includegraphics[width=1\textwidth]{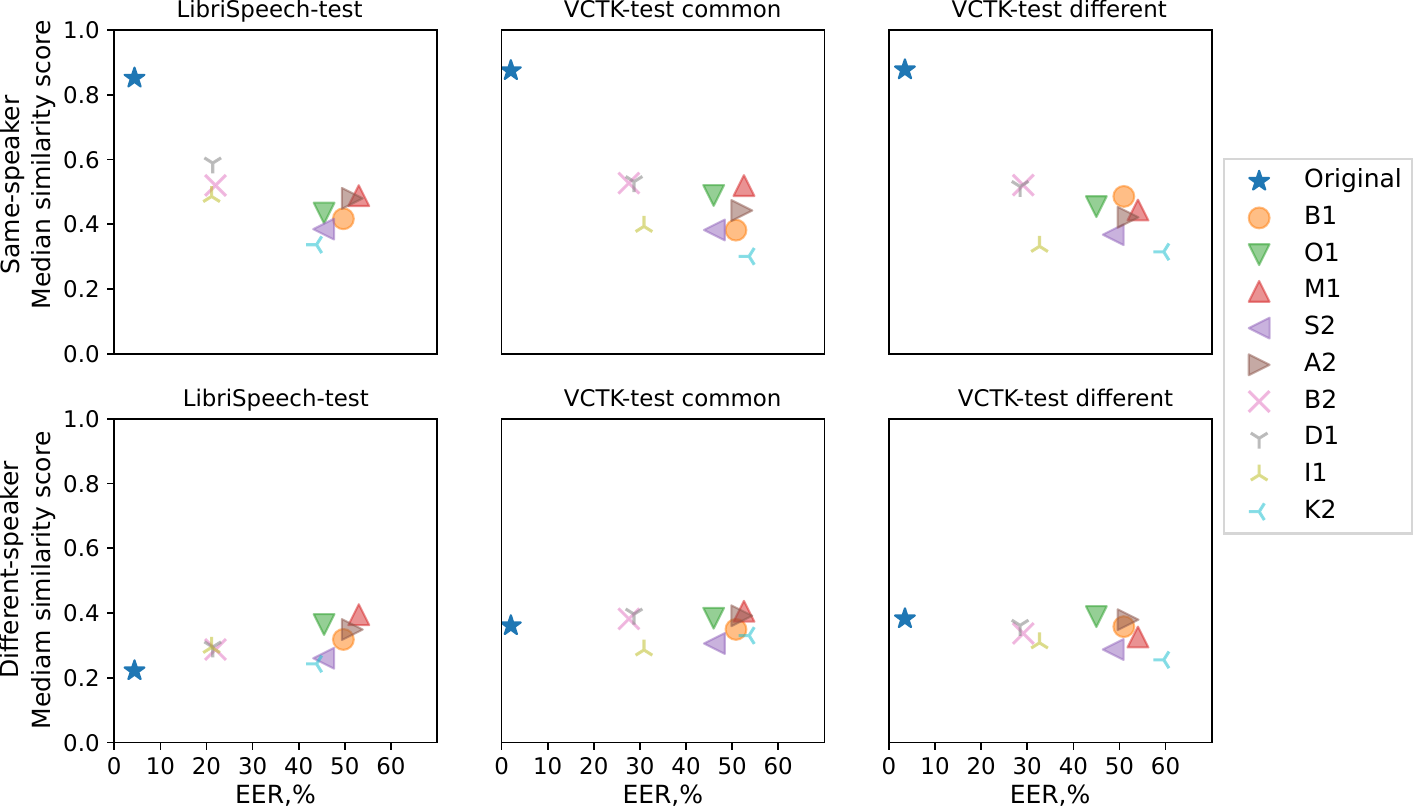}}
\caption{Objective EER \added{(ignorant attack model)} vs.\ subjective \added{same- or different-speaker speaker similarity scores} on {\it LibriSpeech-test} and the two subsets of {\it VCTK-test}.}
\label{fig:obj-vs-subj-eer}
\end{figure}

In this section, we compare objective and subjective evaluation results. Figure~\ref{fig:obj-vs-subj-eer} plots the EER against the \added{median subjective speaker verifiability (similarity) score} for all primary anonymization systems and for original data (blue star) on the three test sets.
The results indicate that anonymizing the trial \added{increases the EER and decreases the same-speaker subjective similarity score, while it leaves the different-speaker similarity score roughly unchanged}. The precise impact depends on the anonymization system and the test set. 
This suggests that the considered anonymization systems can hide the speaker identity to some degree from both ASV system and human ears. This is an encouraging message from the challenge.
Similar results can be observed for other privacy metrics, as shown by 
\citet[Section~5]{Tomashenko2021CSlsupplementay}.

\begin{figure}[h!]
\begin{center}
\center{\includegraphics[width=0.68\textwidth]{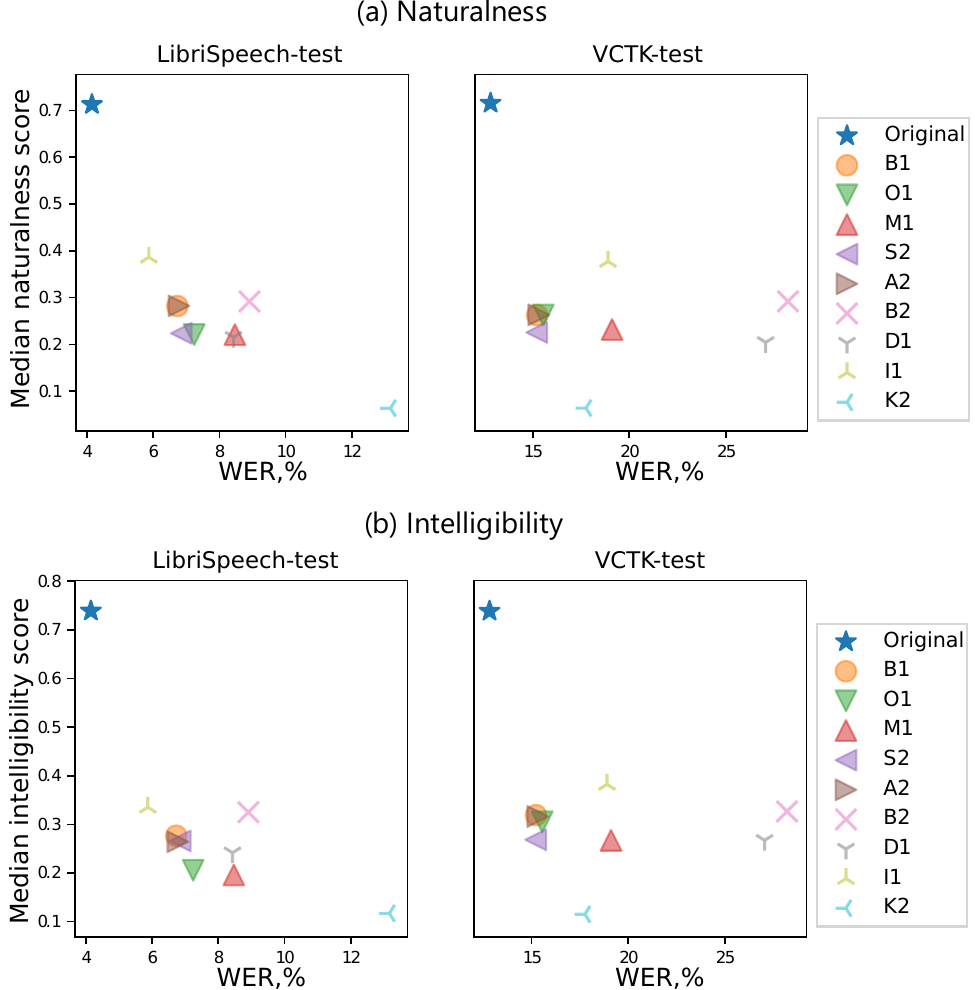}}
\medskip
\end{center}
\caption{Objective WER vs.\ subjective naturalness and intelligibility scores averaged over {\it LibriSpeech-test} and {\it VCTK-test}.}
\label{fig:obj-vs-subj}
\end{figure}

 Figure~\ref{fig:obj-vs-subj} plots the WER against the median subjective naturalness and intelligibility scores, averaged over all test datasets. The results reinforce the observation made earlier that all anonymization systems degrade the objective and subjective utility metrics.  
On \textit{LibriSpeech-test}, the best results for all utility metrics is achieved by the signal processing based system \textbf{I1}, and the worst one by system \textbf{K2}. However, on \textit{VCTK-test}, there is no system that performs best (or worst) for all metrics. This is mostly due to the fact that the WER is less consistent across datasets than the subjective naturalness and intelligibility scores.

\section{Conclusions}\label{sec:conclusions}

The VoicePrivacy 2020 Challenge was conceived to promote private-by-design and private-by-default speech technology and is the first evaluation campaign in voice anonymization.  
The voice anonymization task is defined as a game between users and attackers, with three possible attack models
each corresponding to  adversaries with different knowledge of the applied anonymization methods.
The paper describes a full evaluation framework for the benchmarking of different anonymization solutions, including datasets, experimental protocols and metrics, as well as two open-source baseline anonymization solutions in addition to the comprehensive objective and subjective evaluation of both baseline systems and those submitted by challenge participants.
These indicate the potential for successful anonymization and serve as a platform for future work in what is now a burgeoning research field.

\subsection{Summary and findings}

 The challenge attracted participants from both academia and industry, including experts already working on anonymization and people new to the field.
 The submitted anonymization systems can be broadly classified into two classes: x-vector based systems relying on speech synthesis (such as the primary baseline \textbf{B1}) and signal processing based systems (relating to the secondary baseline \textbf{B2} and system \textbf{I1}).
 X-vector based systems provide the best objective results 
 on average.\footnote{There are some exceptions, related to the WER results for system \textbf{I1} and the \textit{LibriSpeech} dataset in particular.}
 In contrast, subjective evaluation shows that signal processing based systems
 tend to yield higher naturalness and intelligibility.

 More consistent findings show that 
anonymization produced by all systems degrade naturalness and intelligibility, as well as the WER.  
Furthermore, the best systems in terms of WER are based on x-vector anonymization
whereas the best system in terms of intelligibility is system \textbf{I1}.  

Anonymization is also achieved only partially and always at the cost of utility; no single system gives the best performance for all metrics and each system offers a different trade-off between privacy and utility, whether judged objectively or subjectively.
This finding holds irrespective of the attack model.  While for the ignorant attack model, many systems achieve EERs above $50\%$, 
the best results are in the range of $33-43\%$ for the lazy-informed attack model, and in the range of $16-26\%$ for the semi-informed attack model.  System rankings are also different in each case, demonstrating the difficulty of designing an anonymization system that performs well across the range of different VoicePrivacy attacks models.

Challenge participants  investigated the proposed  anonymization approaches and suggested  improvements in some test-cases over the baseline anonymization solutions. They found out, that (1)~resynthesis alone degrades utility, while also improving privacy; (2)~there is potential for privacy leakage not only in x-vector embeddings, but also in phonetic features and pitch estimates \cite{champion2020speaker,Mawalim2020}; (3) the distribution of anonymized x-vectors differs from that of original x-vectors \cite{turner2020speaker}. 
Recent work shows the potential to reduce privacy leakage in pitch estimates while also protecting utility \cite{champion2020astudyf0,srivastava2021}.
Other findings show that degradations to utility can be mitigated by retraining models used for downstream goals, such as ASR, using anonymized data.  Lastly, we identified some differences or bias in performance across different datasets and for different speaker gender.  The scale of these differences is one factor, among others discussed below, that warrants further attention in future research.

\subsection{Open questions and future directions}

A common understanding of VoicePrivacy is still in its infancy. For one, communicating the achievements in layperson terms remains a challenge to better integrate the larger speech community and for outreach to the public at large; for another, VoicePrivacy cannot remain at scratching the surface of privacy issues related to speech and language technology. While considering biometric identity as sensitive information in the first edition, there are other types of sensitive information encoded and transported through speech as a communication medium. Moreover, by constraining the first edition to the operability of speech recognition, linguistic features still allow for extracting biometric characteristics to identify authorship. Depending on the context, the settings of ASV and ASR systems, one might argue that for prompted speech in automated call centers, there is less subjective variability in what is said; let alone, the goal of VoicePrivacy as a community is speech technology as a whole.

Future editions of the VoicePrivacy Challenge will include stronger baseline solutions, possible extensions of the tasks, and re-visited evaluation protocols:

\begin{itemize}
    \item \textit{Improved anonymization methods for  stronger baseline solutions.} For the primary baseline and related approaches,  perspective improvements in x-vector based anonymization include adversarial learning \cite{espinoza2020speaker} and design strategies  based on  speaker space analysis, gender, distance metric,  etc. \cite{srivastava2020baseline,srivastava2021}.
    Sensitive information can be further removed from prosodic and other features, in particular, from pitch \cite{srivastava2021,champion2020astudyf0,gaznepoglu2021exploring} and phonetic (BN)  features.
    Improved algorithms to use the speaker pool should take into account not only speaker characteristics before anonymization but also voice distinctiveness after anonymization.  
    Moreover, the quality of the synthesized speech using unseen x-vectors has room for improvement. 
    For the secondary baseline, we will consider its extension using a stochastic choice of McAdams' coefficient \cite{patino2020speaker}.

    \item \textit{Stronger and more realistic attack models}. Development and investigation of stronger  
    attack models is another potential direction. 
 A knowledgeable and experienced adversary will improve
the ASV system and adapt it to make better decisions, i.e., to yield better
class discrimination alongside accurate forecasts. Contrary to the conventional
experimental validation based on error rates, an adversary actually needs to put
 a specific threshold and might want to change this threshold, depending on the
settings of the ASV systems.
In other words, priors and costs that determine the decision policy of an adversary need
to be highly adaptable.

\item \textit{Alternative privacy and utility metrics \added[id=rev, comment=2.3]{and datasets.}}
The ongoing work on privacy preservation assessment is focusing on the development of  new evaluation frameworks, anonymization metrics, and  investigation of  their correlation and complementarity. This includes the ZEBRA framework \cite{nautsch2020zebra, noe2021csl}, and objective and subjective linkability metrics \cite{maouche2020acomparative}.
Also one may be interested in evaluation that is close to  real industry applications and tasks, for example, speaker labeling for diarization, analysis of time and quality required for annotation of real vs.\ anonymized speech \cite{espinoza2020speakerpresent}.
The metrics considered in the challenge do not evaluate fully the requirement that all  characteristics in the speech signal except the speaker identity should be intact. %
Relevant utility metrics depend on the user's downstream goals, and for additional downstream goals other utility
metrics should be considered. \added[id=rev, comment=2.3]{This will require additional datasets for which these goals have been annotated. Datasets collected in real usage conditions should also be considered to assess the impact of acoustic conditions (reverberation, noise, overlapping speech) and full conversations.}

\item \textit{Attributes}. Besides the speaker identity information, speech also conveys other attributes that can be considered  as sensitive, such as emotional state, age, gender, accent, etc. Selective suppression of such attributes is a possible task extension. \added[id=rev, comment=2.3]{Except for age and gender which are available in \textit{LibriSpeech}, this will require additional datasets for which these attributes have been annotated.}

\item \textit{Privacy vs utility trade-off}.
The privacy is often achieved at the expense of utility, and an important question  is how to set up  a proper threshold between privacy and utility \cite{li2009tradeoff}. 
When developing anonymization methods, a joint optimization of utility gain and privacy loss can be performed by incorporating them into the criterion for training anonymization models \cite{Hiroto2021}.

\item \textit{Integrated approach to voice privacy and security}.
In the bigger picture, security and privacy need to be thought of together and not as opposing forces: positive-sum solutions \cite{Cavoukian2017} need to be sought to design technology for better products and services. In other words, while one might draw inspiration from machine learning, forensic sciences, and biometrics, integrated privacy designs for speech and language technology must sacrifice neither security, business interests, nor privacy.  Developing of adequate VoicePrivacy safeguards demands future directions that empower capacity for their credible and adequate use in integrated privacy designs which beyond technology include organisational measures.
   
\end{itemize}

\section*{Acknowledgment}
VoicePrivacy was born at the crossroads of projects VoicePersonae, COMPRISE (\url{https://www.compriseh2020.eu/}), and DEEP-PRIVACY. Project HARPOCRATES was designed specifically to support it. The authors acknowledge support by ANR, JST (21K17775), and the European Union's Horizon 2020 Research and Innovation Program,  and they would like to thank Christine Meunier.

\bibliography{mybibfile}

\end{document}